\documentclass[sigconf, screen]{meta/acmart}
\usepackage{multirow}
\usepackage{pifont}
\usepackage{makecell}
\usepackage{xspace}
\usepackage{wasysym}
\usepackage{arydshln}
\usepackage{svg}
\newcommand{\nop}[1]{}
\newcommand{\edge}{\textit{EDGE}\xspace}
\renewcommand\footnotetextcopyrightpermission[1]{} % removes footnote with conference information in first column
\definecolor{mycustompurple}{RGB}{154, 36, 79} % 定义自己的颜色
\usepackage[utf8]{inputenc}
\AtBeginDocument{%
  }

\acmConference[Under Review]{Under Review}{2024}{}
\begin{document}

\title{EDGE: Enhanced Grounded GUI Understanding with Enriched
Multi-Granularity Synthetic Data}

\author{Xuetian Chen}
\affiliation{%
  \institution{Fudan University}
  \city{Yangpu Qu}
  \state{Shanghai Shi}
  \country{China}}
\email{23210980105@m.fudan.edu.cn}

\author{Hangcheng Li}
\affiliation{%
  \institution{Fudan University}
  \city{Yangpu Qu}
  \state{Shanghai Shi}
  \country{China}}
\email{22307130449@m.fudan.edu.cn}

\author{Jiaqing Liang}
\affiliation{%
  \institution{Fudan University}
  \city{Yangpu Qu}
  \state{Shanghai Shi}
  \country{China}}
\email{liangjiaqing@fudan.edu.cn}

\author{Sihang Jiang}
\affiliation{%
  \institution{Fudan University}
  \city{Yangpu Qu}
  \state{Shanghai Shi}
  \country{China}}
\email{jiangsihang@fudan.edu.cn}

\author{Deqing Yang}
\affiliation{%
  \institution{Fudan University}
  \city{Yangpu Qu}
  \state{Shanghai Shi}
  \country{China}}
\email{yangdeqing@fudan.edu.cn}

\renewcommand{\shortauthors}{Xuetian, et al.}

\setcopyright{none}

\begin{abstract}
Autonomous agents operating on the graphical user interfaces (GUIs) of various applications hold immense practical value. Unlike the large language model (LLM)-based methods which rely on structured texts and customized backends, the approaches using large vision-language models (LVLMs) are more intuitive and adaptable as they can visually perceive and directly interact with screens, making them indispensable in general scenarios without text metadata and tailored backends. Given the lack of high-quality training data for GUI-related tasks in existing work, this paper aims to enhance the GUI understanding and interacting capabilities of LVLMs through a data-driven approach. We propose \edge, a general data synthesis framework that automatically generates large-scale, multi-granularity training data from webpages across the Web. Evaluation results on various GUI and agent benchmarks demonstrate that the model trained with the dataset generated through \edge exhibits superior webpage understanding capabilities, which can then be easily transferred to previously unseen desktop and mobile environments. Our approach significantly reduces the dependence on manual annotations, empowering researchers to harness the vast public resources available on the Web to advance their work. Our source code, the dataset and the model are available at \url{https://anonymous.4open.science/r/EDGE-1CDB}.
\end{abstract}

\nop{

}




\maketitle

\section{Introduction}

Autonomous interaction with computing devices has long been a topic of artificial intelligence research \cite{bolt1980put, lieberman1995letizia}. With the popularity of personal computers and smartphones, the agents that can automatically interact with graphical user interfaces (GUIs) of various applications have become a growing focus \cite{shi2017world, li2020mapping, he2021actionbert, bai2021uibert}. Recently, the continuous advance of large language models (LLMs) \cite{achiam2023gpt, anthropic2023claude, team2023gemini} makes it increasingly feasible to develop general GUI agents. 

\begin{figure}[h]
\centering
\includegraphics[width=0.85\linewidth]{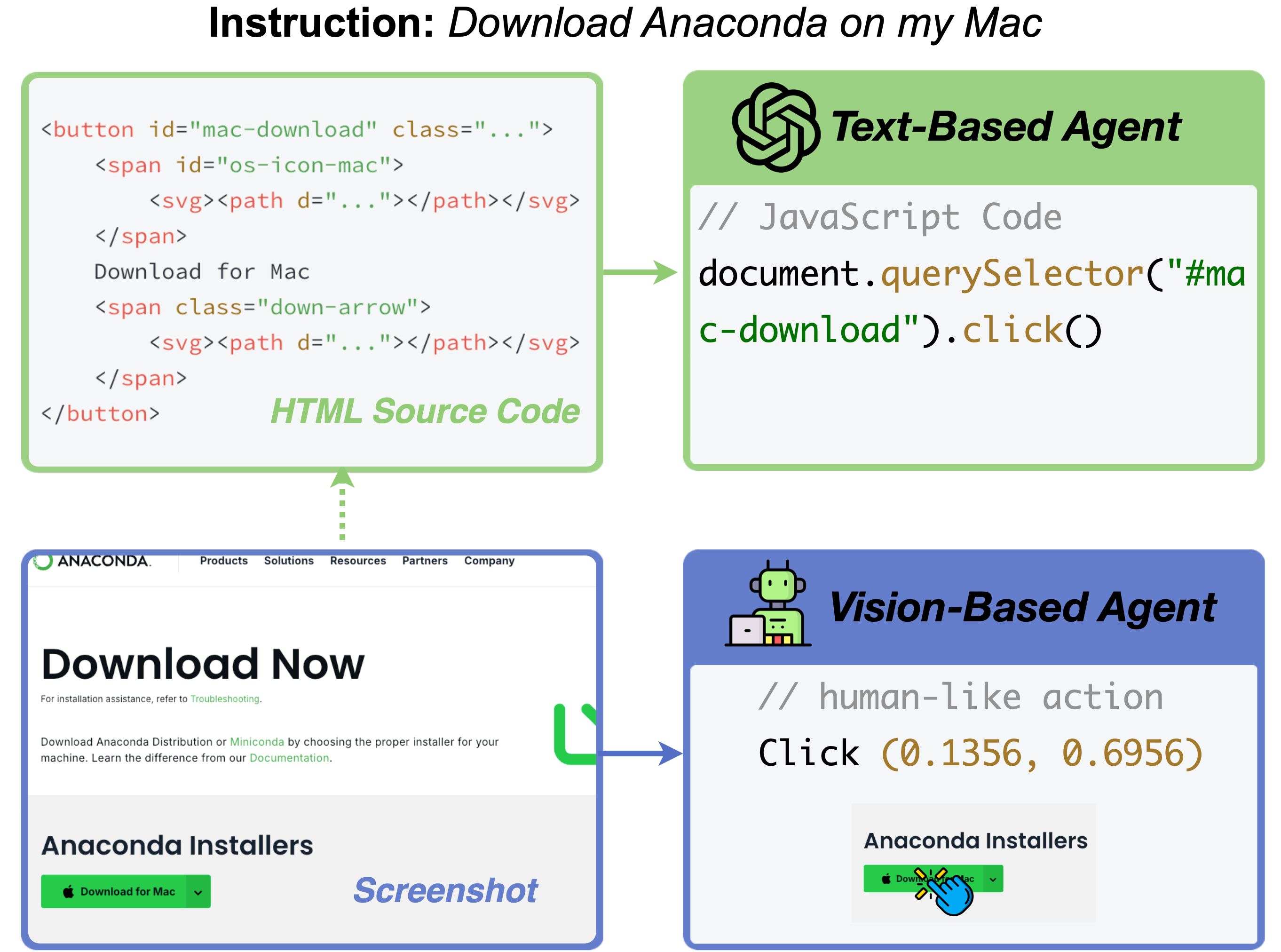}
\vspace{-0.1cm}
\caption{Text-based agents take extracted textual metadata as input (e.g., HTML) and perform actions through specific backends (such as browser engines). Vision-based agents directly read the screen and execute actions like mouse clicks.}
\Description{Differences between input and output of text-based agents and vision-based methods.}
\label{fig:visual}
\vspace{-0.4cm}
\end{figure}

Numerous studies \cite{gur2023real, rawles2024androidinthewild, wen2024autodroidllmpoweredtaskautomation, deng2024mind2web, lai2024autowebglm, Significant_Gravitas_AutoGPT, GPT-4V-Act, wu2024copilot, gao2023assistgui} have proposed LLM-based GUI agents on web, mobile, and desktop environments. These methods avoid visually perceiving GUIs by using structured texts (e.g., HTML for webpages \cite{deng2024mind2web, gur2023real} and view hierarchy for Android screens \cite{rawles2024androidinthewild, wen2024autodroidllmpoweredtaskautomation}) as input. In addition, they may require environment-specific backend access to control the systems \cite{Significant_Gravitas_AutoGPT, wu2024copilot}. However, given that GUIs are designed for human interaction through visual reading and basic actions like clicking, the methods relying on text metadata or customized backend are neither intuitive nor widely practical. Since GUI formats vary across different environments, the agents in generalized scenarios must interact with screens in a human-like manner, without access to underlying data. 

Recognizing this limitation, recent studies \cite{li2022spotlight, shaw2023pixels, yan2023gpt, hong2023cogagent, yang2023appagent, niu2024screenagent, cheng2024seeclick, you2024ferret} have turned to the vision-based approaches that employ large vision-language models (LVLMs) to directly perceive and interact with screens, as illustrated in Figure~\ref{fig:visual}. To improve the limited capabilities of LVLMs in grounded GUI understanding and interacting, where "grounded" refers to the ability to precisely locate elements, these efforts often involve further training on the GUI-specific datasets with element-level annotations, which are typically sourced from the Web \cite{hong2023cogagent, cheng2024seeclick, you2024ferret, gao2024enhancing}. Specifically, they drive browser engines to capture the element-level correspondences between HTML and rendered webpage images, including both positions and content. This automated process exploits the supervisory signals inherent on Web, reducing the need for manual annotations.

However, models trained using these methods still exhibit some drawbacks in real-world scenarios. Firstly, these models generally underperform on the GUIs with rich visual elements such as images and icons. Secondly, despite their fine performance on element-level grounded understanding, they are not competent on the action grounding tasks, where the models are required to map real instructions to certain operations on target elements. Thirdly, these models have difficulty understanding the local and global semantics that humans can easily capture, resulting in possible failure on tasks beyond the single-element level.

Therefore, following a data-driven approach, we propose \edge, an automated webpage annotation and data synthesis framework. \edge is characterized by its rich annotations and multi-granularity task settings, aimed at enhancing the capabilities of open-source LVLMs in grounded GUI understanding and interaction, while also promoting further research on general GUI agents. Specifically, in the annotation stage, we extract explicit texts as well as rich latent elements from webpages, with an additional collection of extensive icons commonly used in GUIs to address the challenges of icon understanding. Subsequently, in addition to the basic element-level recognition question-answering (QA), \edge also integrates the multi-granularity QA tasks synthesized using the powerful proprietary model Claude-3.5 \cite{anthropic2023claude}. This comprehensive task setup is designed to cover the global semantic understanding and reasoning abilities required in complex webpage and action grounding scenarios. Besides, through our framework, the model's GUI-related capabilities learned from large-scale webpages can be easily transferred to desktop and mobile environments, which lack available supervised data and typically rely on manual annotations.

Our main contributions are summarized as follows:
\begin{itemize}
    \item We propose \edge, an automated webpage annotation and data synthesis framework that significantly enriches annotations and covers multi-granularity tasks to enhance the GUI-specific capabilities of LVLMs.
    \item Following \edge, we constructed a large-scale dataset of GUI tasks and release it together with the LVLM trained on it, which has excellent grounded GUI understanding and interacting capabilities.
    \item The model's outstanding performance on GUI benchmarks \cite{cheng2024seeclick, liu2024visualwebbench} and the gains achieved on downstream agent benchmarks \cite{shi2017world, rawles2024androidinthewild, deng2024mind2web} demonstrate the effectiveness of \edge.
\end{itemize}

\section{Related Work}
    \subsection{Large Vision-Language Models}
        Various studies have proposed solutions to address the challenges \cite{yin2023survey, song2023bridge, zhang2024mm} of architectural complexity and data scarcity in the training of LVLMs. A series of methods \cite{alayrac2022flamingo, li2023blip, li2023blip, ye2024mplug, wang2023cogvlm} utilize elaborately designed architecture to integrate visual encoders with LLMs, thereby inheriting the language and reasoning abilities of LLMs. For data scarcity, the LLaVA series \cite{liu2024visual, liu2024improved} utilize GPT-4 \cite{achiam2023gpt} to generate rich instruction-following data. Monkey \cite{li2024monkey} proposes a multi-level caption generation pipeline. Such data-driven approaches are more common in training domain-specific knowledge, such as medicine \cite{li2024llava}, and spatial awareness \cite{chen2023shikra, zhang2023gpt4roi, peng2023kosmos}.

        In this work, we focus on GUI environments and adopt the data-driven approach to address the limitations in both the quantity and diversity of training data.

    \subsection{GUI Understanding and Interacting}
        Previous work primarily targets simplified web \cite{shi2017world, liu2018reinforcement, gur2018learning, lee2023pix2struct, shaw2023pixels} or mobile environments \cite{li2020mapping, burns2022dataset, li2022spotlight}. 
        Recent emergence of real-world benchmarks \cite{deng2024mind2web, zhou2023webarena, rawles2024androidinthewild} promote an LLM-centric paradigm. Several methods explore prompting GPT-4 for web tasks via in-context learning \cite {zheng2023synapse} and self-refinement \cite{kim2024language}. WebAgent \cite{gur2023real} enhances manipulation through instruction decomposition and programming.

        Recently, research focus has shifted to more intuitive and adaptable LVLM-based methods. Mobile-Agent \cite{wang2024mobile} uses external detection models to determine the positions of interactive components. A series of studies \cite{yan2023gpt, gao2023assistgui, yang2023appagent, zheng2024gpt} rely on the powerful visual perception ability of GPT-4V, combined with extracted text metadata to compensate for its lack of grounding capabilities \cite{achiam2023gpt}. 
        
        Latest work like CogAgent \cite{hong2023cogagent}, SeeClick \cite{cheng2024seeclick}, and Ferret-UI \cite{you2024ferret} neither depend on proprietary models nor GUI metadata. Our work, following the same motivation, aims to empower open-source LVLMs with GUI understanding and interacting capabilities.
        
    \subsection{Grounding Capability of LVLMs}
        Researchers have proposed various improvements to the grounding performance of LVLMs, including introducing external vision models \cite{zhao2023bubogpt, wang2024mobile}, internally integrating fine-grained visual modules \cite{zhang2023llava, you2023ferret, lai2024lisa}, and directly predicting bounding boxes \cite{chen2023shikra, peng2023kosmos, zhang2023gpt4roi}.

        Classical detection models \cite{carion2020end, zhu2020deformable, liu2023grounding} are not well-suited for GUI environments due to semantic disparities, resulting in a scarcity of detailed annotated datasets. CogAgent \cite{hong2023cogagent} leverages crawled webpages to collect large-scale annotations, supplemented by manually annotated fine-grained data. SeeClick \cite{cheng2024seeclick} develop a simple framework for annotation grounding dataset synthesis. Ferret-UI \cite{you2024ferret} uses the UI detection model to annotate mobile screens, as well as to synthesize elementary and advanced tasks by GPT-4.

        Inheriting the advantages of these works, we aim to generate richer annotations from public webpages and automatically synthesize multi-granularity data with the aid of proprietary models.

\section{Preliminaries}
    \subsection{Motivation}
        In humans' view, understanding and interacting with a GUI requires perceiving the content and locations of on-screen elements and accurately positioning the target element. Based on this intuition, we believe that the poor performance of existing general-purpose LVLMs \cite{achiam2023gpt, anthropic2023claude, team2023gemini, li2024llava, bai2023qwenvl} in GUI tasks \cite{cheng2024seeclick, liu2024visualwebbench} stems from their lack of grounded GUI understanding and interacting capabilities.
        
        On one hand, GUIs are characterized by flat, dense elements and a mix of text and icons, hinder the transfer of some models' built-in grounding abilities in natural images \cite{bai2023qwenvl} to GUI environments. On the other hand, even though proprietary models \cite{achiam2023gpt, anthropic2023claude, team2023gemini} possess universal image understanding capabilities, the inherent weakness in object detection prevents them from locating elements on the screen. Therefore, models must undergo element-level grounded training within GUI environments to enhance their performance.
    
        \edge is designed to address the lack of large-scale, high-quality datasets in this domain. Before describing the synthesis framework and data pipeline, we first formalize the definition of GUI grounding capabilities of LVLMs and its training process.
    
    \subsection{GUI Grounding Tasks}  \label{sec:grounding}
        We first introduce \textbf{element grounding} and \textbf{action grounding} tasks. Similar to open-set object detection \cite{liu2023grounding}, element grounding requires models to find the location $l$, as coordinates of a point $(x, y)$ or a bounding box $(x_1, y_1, x_2, y_2)$, of an element within the input screenshot $s$ based on a text description $t$, i.e., to predict $p(l~|~s, t)$. Action grounding shares the same input-output format but here $t$ refers to a user instruction. The model must identify the target element before locating it, as the instruction may differ from the text of the target element. The reverse of grounding, namely predicting $p(t~|~s, l)$, is called \textbf{referring}. Grounding and referring tasks often appear in pairs in \edge, mutually reinforcing each other to promote cross-modal semantic fusion.
        
        Unlike vision models \cite{carion2020end, zhu2020deformable, liu2023grounding} that directly predict numeric coordinates, LVLMs typically treat coordinates as part of the output tokens. Previous studies \cite{wang2021screen2words, shaw2023pixels} introduce extended vocabularies (such as $\{\text{<}p_0\text{>}, \text{<}p_1\text{>}, \dots, \text{<}p_{999}\text{>}\}$) to encode coordinates, whereas we follow a more intuitive manner that directly treats them as native language tokens \cite{chen2023shikra, bai2023qwenvl}. This method eliminates the need for additional pre/post-processing without compromising performance. In this setting, training completely follows the standard next-token prediction with cross-entropy supervision.
    
\section{Methodology}
    \subsection{Overview}

    Following the human intuition, the overall task settings in \edge primarily encompass the \textbf{elementary tasks} and \textbf{advanced tasks} as named by Ferret-UI \cite{you2024ferret}. The elementary tasks are designed to enhance the element-level grounded understanding of the GUI, while the advanced tasks introduce the multi-granularity understanding and reasoning capabilities required for GUI interactions.
    
    The training of GUI grounding requires massive element-level annotations. Following previous work \cite{hong2023cogagent, cheng2024seeclick, gao2024enhancing}, we collect general webpages from Common Crawl\footnote{\url{https://commoncrawl.org}}, a large-scale web crawl data repository, and drive Playwright\footnote{\url{https://playwright.dev}} to automatically annotate them. Then we format these annotations to synthesize question-answering (QA) data using various templates. This process is detailed in Section~\ref{sec:elem_tasks}.

    In terms of advanced tasks, We additionally collect highly interactive mainstream websites and leverage proprietary models \cite{anthropic2023claude} to synthesize more flexible and interaction-focused QA pairs, along with detailed descriptions that assess multi-granularity understanding capabilities.  We introduce this process in Section~\ref{sec:adv_tasks}.

    Considering the weaknesses of existing open-source LVLMs, we additionally propose the \textbf{icon understanding task}. This task utilizes an automatically collected icon annotations dataset to enhance the model's ability to understand and interact with icons, as outlined in Section~\ref{sec:icons}.

    % Through the training on these advanced tasks, the model becomes capable of understanding user intention and correctly responding to interaction-related queries regarding the GUI, laying a solid foundation for downstream agent tasks.
    
    \subsection{Elementary Tasks} \label{sec:elem_tasks}
        The elementary tasks primarily synthesize element-level QA from large-scale general web annotations to equip the model with grounded GUI understanding, thereby promoting effective interactions. We introduce the generation in three stages: collection, annotation, and synthesis.
    
        \subsubsection{Collection}
            Due to the vast quantity and varying quality of webpages on Common Crawl, instead of performing our own cleaning, we collect webpages from a refined subset FineWeb-Edu \cite{lozhkov2024fineweb-edu}. FineWeb-Edu consists of carefully cleaned and quality-assessed pages, retaining those that are most educationally valuable for LLMs and typically exhibit well-designed layouts in their GUIs.
    
        \subsubsection{Annotation}
\begin{figure}[h]
            \centering
            \includegraphics[width=0.95\linewidth]{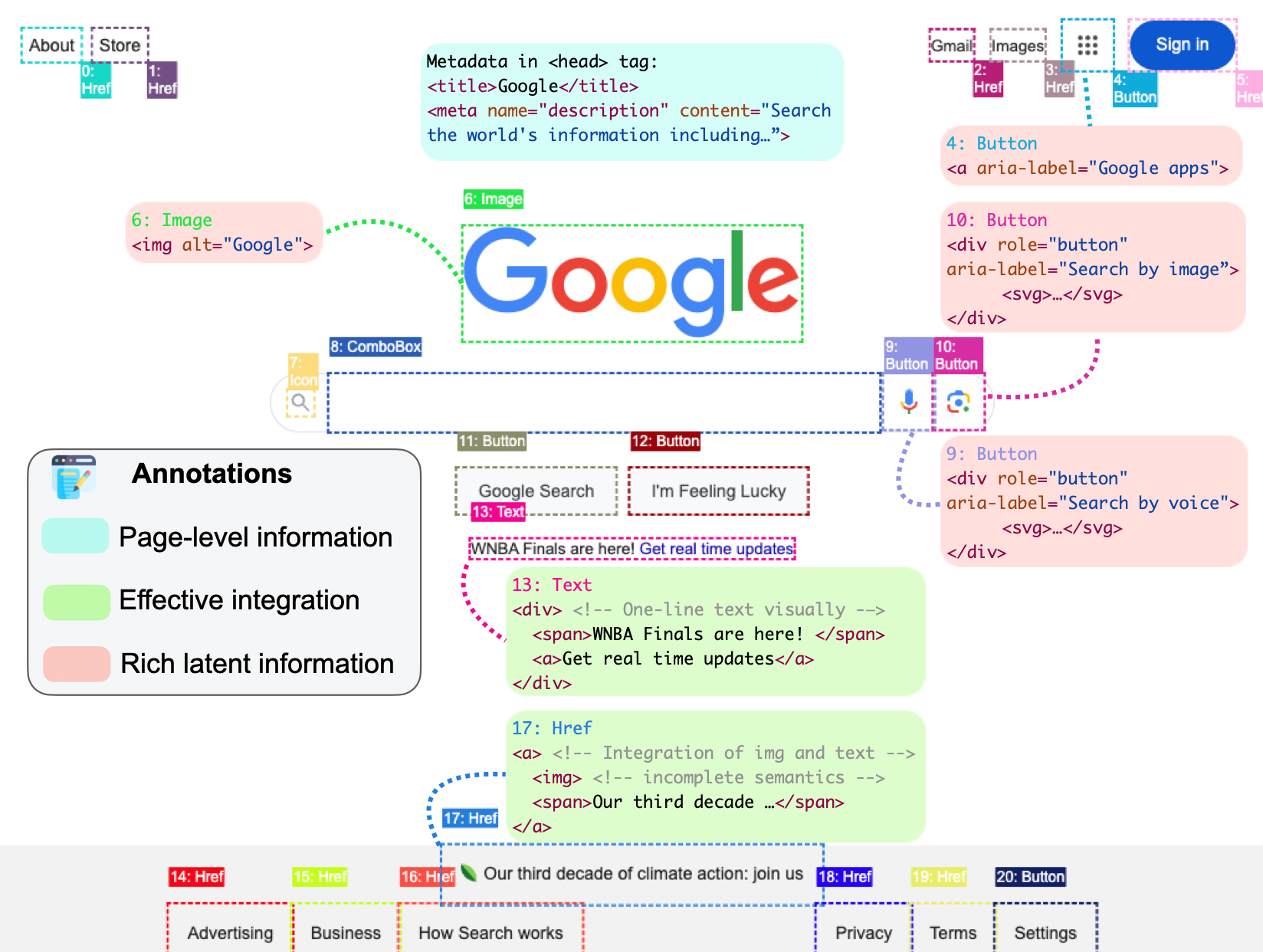}
            \captionsetup{skip=3pt}
            \caption{The diagram of the annotation stage, where the extraction of rich latent semantics and the information integration process is highlighted with the (simplified) HTML of the relevant elements presented. Elements are marked with rectangular boxes for demonstration purposes only.}
            \Description{The diagram of the annotation process, showing the annotation results on Google's homepage and highlighting the characteristics of the \edge method.}
            \label{fig:annotations}
            \end{figure}

            In this stage, we render the webpages, take screenshots and capture positions and content of on-screen elements by injecting JavaScript scripts. Compared to previous work \cite{hong2023cogagent, GPT-4V-Act, cheng2024seeclick, gao2024enhancing}, our annotations introduce improvements in both the effectiveness and richness, making them closer to human-generated ones. Figure~\ref{fig:annotations} showcases the characteristics of our method with the annotation results on the Google homepage.

            \paragraph{Effectiveness}
                Given that there is no unified design standard at the source code level across different websites, identifying the minimal semantic units is crucial for effective annotation. Our carefully designed script includes steps for visibility filtering, classification and integration, ensuring that annotations align with visual presentation and human cognitive habits. Technically, instead of focusing solely on leaf nodes of the DOM tree, we adopt a set of rules to integrate the minimal semantic units based on the structure and visual representation of common tags (e.g., \texttt{<a>} and \texttt{<button>}). More technical details of the script are presented in the Appendix \ref{app:script}.

            \paragraph{Richness}
                Besides explicit text, we also capture the various latent description, such as \texttt{alt} (\texttt{<img>}), \texttt{title} and accessibility content (\texttt{aria-label}). These properties are designed for assistive technologies and are highly suitable as training data for GUI understanding tasks, yet existing annotation methods \cite{GPT-4V-Act, cheng2024seeclick} tend to ignore them.
    
        \subsubsection{Synthesis}
            Based on the annotations, we synthesize element-level grounding and OCR tasks, including elements with explicit text and accessibility labels. We adopt the multi-turn conversation format, where a training sample contains multiple QAs on one screenshot. We additionally synthesize the basic page-level QA about the title and description of the webpage. This process is illustrated in the top half of Figure~\ref{fig:tasks} and the final form of training samples are shown in the Appendix \ref{app:dataset}.

            In our early explorations, the model could roughly locate elements but lacked precision. Hence, we designed two data augmentation techniques to address this weakness:
            
            1. We randomly crop the screenshots, accordingly modifying the coordinates of visible elements. The model is compelled to identify subtle positional variations from similar image features.

            2. We overlay a rectangular box on a randomly selected element within the screenshot. This new task requires the model to simultaneously identify the both position and content of the "highlighted" element, promoting the fusion across modalities.

    \begin{figure*}
        \centering
      \includegraphics[width=0.85\textwidth]{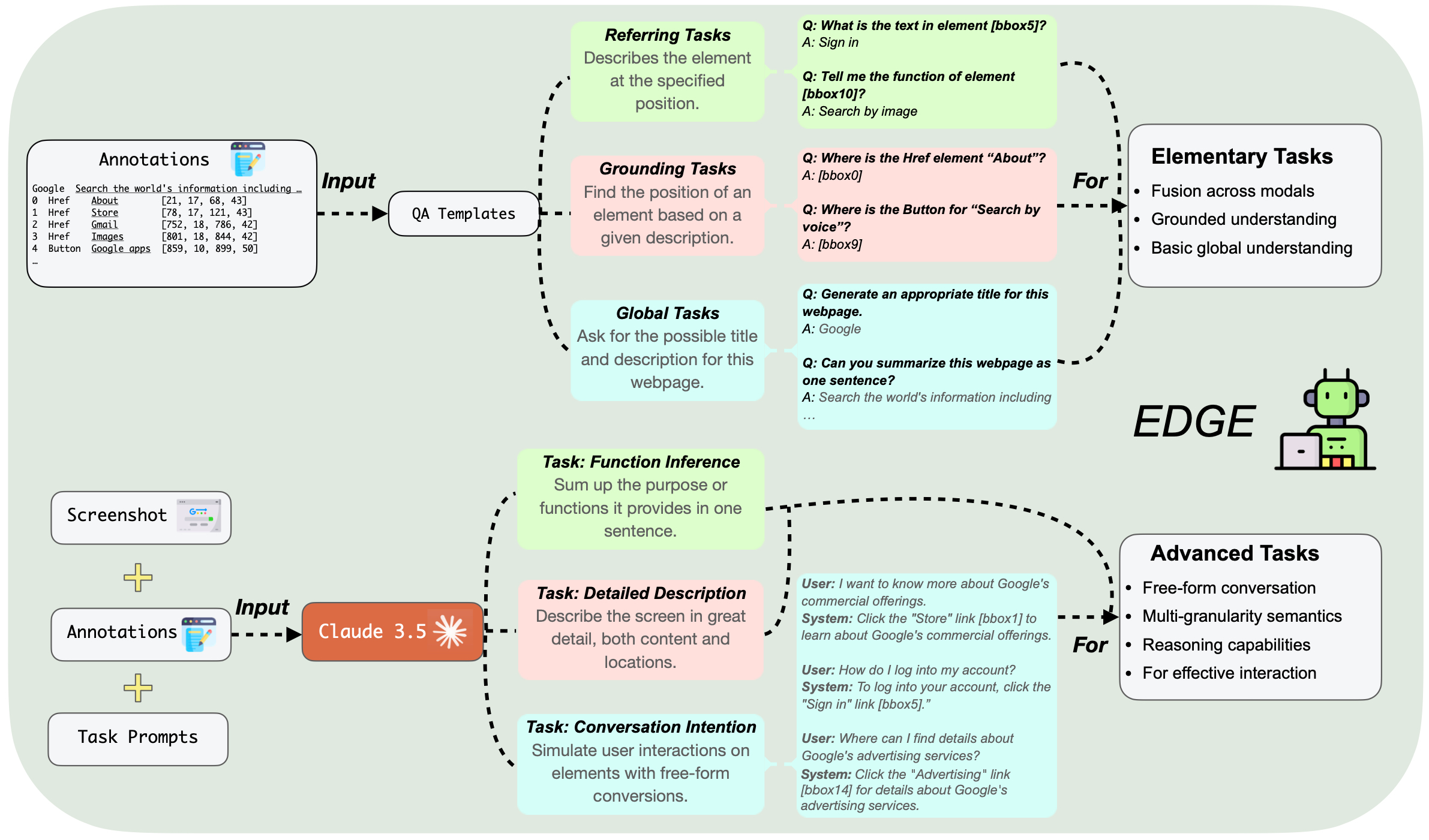}
        \captionsetup{skip=3pt}
        \caption{The synthesis of the elementary and advanced tasks.}
        \Description{A diagram describing the process of generating primary and advanced tasks, also using the Google homepage as an example.}
        \captionsetup{skip=5pt}
        \label{fig:tasks}
    \end{figure*}

    \subsection{Advanced Tasks} \label{sec:adv_tasks}
    Unlike the elementary tasks which focus on understanding through template-based generation, the advanced tasks aim to incorporate the multi-granular understanding and reasoning abilities to respond to free-form interactive instructions. Therefore, we utilize the same annotation methodology to generate high-quality QA pairs on a set of highly interactive mainstream websites, with the help of Claude-3.5 \cite{anthropic2023claude}. Through the training on these advanced tasks, the model becomes capable of understanding user intention and correctly responding to interaction-related queries regarding the GUI, laying a solid foundation for downstream agent tasks.
        \subsubsection{Collection}
            To facilitate GUI interaction, we need more interactive webpages, rather than the aforementioned general ones which are primarily information-focused such as news and blogs. Hence, we collected the top 20,000 domains from Ahrefs Rank\footnote{\url{https://app.ahrefs.com/ahrefs-top}}, considering these as the most frequently used websites which are typically highly interactive. 

        \subsubsection{Annotation and Synthesis}
            The homepages of these sites are annotated using the same method as proposed. During the synthesis stage, we transfer the following three advanced tasks settings from the mobile dataset of Ferret-UI \cite{you2024ferret} to our web environment:
            \begin{itemize}
                \item Function inference: Provide a brief summary of the purpose of the webpage or the function it offers to users.
                \item Detailed description: Offer a detailed account of the position and content of each major element on the webpage.
                \item Conversation intention: Select elements and simulate user interactions with them, presented in a QA format.
            \end{itemize}
            
            Among these, function inference requires global semantics perception, detailed description involves a mix of element-level and local-level understanding, while the last one is similar to the action grounding introduced in Section \ref{sec:grounding}, testing reasoning capabilities. 
            
            We input the annotations of webpages into Claude-3.5 \cite{anthropic2023claude} and prompt it to generate QA pairs for these tasks. Considering the complexity of webpages, we also input the screenshots annotated by the Set-of-Mark\cite{yang2023set} prompting method, aiming to enable Claude-3.5 to capture semantics that automatic annotation might miss. The synthesis of advanced tasks is shown in the bottom half of Figure~\ref{fig:tasks}.
            
    \subsection{Icon Understanding} \label{sec:icons}

        Scripts-based annotations struggle to capture the semantics of visual elements without textual content. Since icon understanding is a weakness in most open-source LVLMs \cite{cheng2024seeclick}, \edge also incorporates an automated icon annotating process, adhering to our principle of minimizing manual work.
        
        First, we collect frequently-used icons from several icon font libraries\footnote{\url{https://fontawesome.com/icons}}\footnote{\url{https://getbootstrap.com}}\footnote{\url{https://www.iconfont.cn}}. Then GPT-4 is prompted to generate descriptions for them. The icon-description pairs are directly added to the QA training set. In addition, to simulate the mixed layout of icons with other content in real GUIs, icons are also randomly embedded into webpage screenshots to construct referring and grounding tasks.
        
        Unlike OCR, icon understanding relies more on the similar cases seen during model training. Our collection includes commonly used icons and logos of well-known products, which can effectively address the model's cognitive gap regarding icon-dense GUIs and logos frequently encountered in real-world scenarios.
    
    \subsection{Dataset Statistics} \label{sec:stat}
    \begin{figure}[h]
      \centering
      \includegraphics[width=\linewidth]{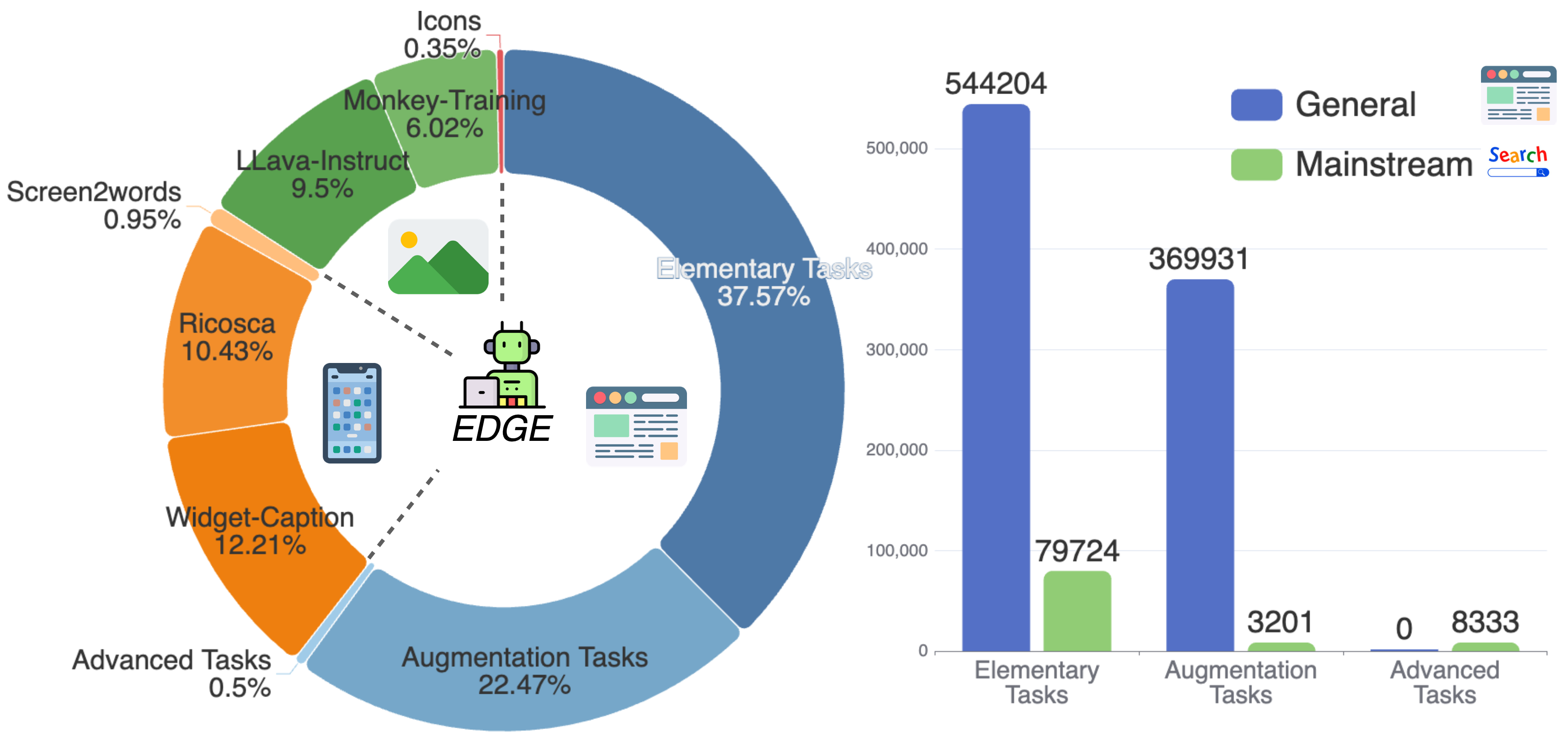}
      \caption{Statistics of \edge with respect to the number of images. The pie chart gives an overview of the distribution of samples from different tasks and environments, while the right displays the number of samples from different webpage sources within three task settings of the web environment.}
      \Description{The pie chart of the \edge dataset shows the various task components of the dataset.}
      \label{fig:dataset}
    \end{figure}

    To facilitate the transfer of web GUI capabilities to general GUI environments, while retaining the ability to comprehend natural images, \edge also includes three tasks derived from the mobile GUI annotation dataset Rico \cite{deka2017rico}: Screen2words \cite{wang2021screen2words}, Widget-Caption \cite{li2020widget}, and Ricosca \cite{li2020mapping}, along with general visual instruction-following and QA dataset from the training set of LLaVA \cite{zhang2023llava} and Monkey \cite{li2024monkey}. Together with the webpage dataset synthesized by \edge itself, this forms a comprehensive dataset comprising 5.42 million QA pairs across 1.66 million images, still referred to as \edge for convenience. Approximately 60\% of the images and 85\% of the QA pairs originate from the \edge proprietary synthesis method. Figure~\ref{fig:dataset} illustrates the distribution of task types within the final dataset and the number of samples from different webpage sources in the web environment QAs.

\section{Experiments}
    To validate the effectiveness of \edge, we train a model using it (hereinafter referred to as \edge as well) and compare its performance with some representative LVLMs on GUI-related tasks. We outline the training details in Sec~\ref{sec:training_details}. Then the evaluation results on two GUI-specific benchmarks \cite{liu2024visualwebbench, cheng2024seeclick} and three downstream agent benchmarks \cite{shi2017world, rawles2024androidinthewild, deng2024mind2web} are presented in Sec~\ref{sec:gui_benchmarks} and \ref{sec:agent_benchmarks}, respectively. Finally we introduce the ablation study in Sec~\ref{sec:abl_study}.

    \subsection{Training Details} \label{sec:training_details}
        \edge undergoes further training on the robust baseline, Monkey \cite{li2024monkey}. We freeze the image encoder and train all parameters of the LLM. 
        For accelerating and memory saving, we adopt a Lion 8-bit optimizer \cite{chen2024symbolic} with a learning rate of 1e-5 and a weight decay ratio of 0.1. The cosine learning rate scheduler is employed with a warm-up ratio of 0.02 and the global batch size is 512. Other settings follow Monkey's. Training for one epoch on the full dataset takes approximately 2 days on 8 NVIDIA A800 GPUs.

    \subsection{Results on GUI Benchmarks} \label{sec:gui_benchmarks}
        We evaluated the grounded GUI capabilities of \edge on two recently proposed benchmarks \cite{liu2024visualwebbench, cheng2024seeclick}. In these settings, each sample represents an independent question or instruction, and the model is required to answer the question or locate the target element based solely on the image, without any context.

        \subsubsection{VisualWebBench \cite{liu2024visualwebbench}} \label{exp:vwb}
            Comprising over 1,500 samples from 139 websites, VisualWebBench is designed to comprehensively evaluate the understanding and interaction capabilities of LVLMs on webpages. There are seven tasks of different granularities, ranging from deterministic, elementary OCR and grounding to open-ended captioning that requires global semantics.

            \begin{table*}[ht!] 
                \centering
                \caption{Results on VisualWebBench, which are sourced from \cite{liu2024visualwebbench}, except Monkey, Qwen2-VL and \edge. $\CIRCLE$ and $\Circle$ represent whether the model is GUI-specific or not, and $\RIGHTcircle$ represents a general model that specifically trained with GUI grounding data. For $\CIRCLE$ and $\RIGHTcircle$, results of Element Ground and Action Ground are click accuracy, rather than multi-choice accuracy for $\Circle$. }
                \label{tab:visualwebbench}
                \vspace{-0.2cm}
                    \begin{tabular}{ccccccccccc}
                    \toprule
                    \multirow{2}{*}{Model} & \multirow{2}{*}{\parbox{1.5cm}{\centering Model\\Size}} & \multirow{2}{*}{\parbox{1.5cm}{\centering GUI\\Specific}} & \multicolumn{3}{c}{Website} & \multicolumn{2}{c}{Element} & \multicolumn{2}{c}{Action} & \multirow{2}{*}{Average} \\ \cline{4-10}
                    & & & Caption & WebQA & HeadOCR & OCR & Ground & Prediction & Ground & \\
                    \midrule
                    Monkey~\cite{li2024monkey} & 9.6B & \Circle & 3.2 & 30.0 & 9.6 & 4.7 & 18.4 & 6.4 & 22.3 & 13.5 \\
                    Qwen-VL~\cite{bai2023qwenvl} & 9.6B & \Circle & 21.8 & 32.2 & 48.4 & 13.4 & 14.0 & 26.7 & 10.7 & 23.9 \\
                    CogVLM~\cite{wang2023cogvlm} & 17B & \Circle & 16.6 & 30.6 & 65.9 & 10.0 & 17.7 & 11.7 & 23.3 & 25.1 \\
                    LLaVA-1.6-7B~\cite{liu2024llava} & 7B & \Circle & 27.0 & 39.8 & 57.3 & 54.8 & 31.7 & 30.6 & 10.7 & 36.0 \\
                    LLaVA-1.6-34B~\cite{liu2024llava} & 34B & \Circle & 24.3 & 48.2 & 67.1 & 71.9 & 43.1 & \textbf{74.0} & 25.2 & 50.5 \\
                    Qwen2-VL~\cite{wang2024qwen2} & 8.3B & \RIGHTcircle & 20.4 & 75.5 & 73.4 & 82.1 & 62.0 & 6.8 & 49.5 & 52.8 \\
                    Gemini Pro~\cite{team2023gemini} & - & \Circle & 25.0 & 55.5 & \textbf{75.1} & 65.4 & 44.3 & 26.7 & 43.7 & 48.0 \\
                    Claude Sonnet~\cite{anthropic2023claude} & - & \Circle & 28.9 & \textbf{81.8} & 70.3 & \textbf{89.2} & 68.8 & 63.4 & 58.3 & \textbf{65.8} \\
                    GPT-4V~\cite{achiam2023gpt} & - & \Circle & \textbf{34.5} & 75.0 & 68.8 & 62.8 & 67.5 & 67.6 & \textbf{75.7} & 64.6 \\
                    SeeClick~\cite{cheng2024seeclick} & 9.6B & \CIRCLE & 0.0 & 19.6 & 34.8 & 0.0 & 70.0 & 1.8 & 42.6 & 24.1 \\
                    CogAgent-Chat~\cite{hong2023cogagent} & 18B & \CIRCLE & 16.3 & 53.3 & 20.2 & 32.4 & 46.3 & 13.5 & 58.4 & 34.3 \\
                    \hline
                    \edge & 9.6B & \CIRCLE & 23.8 & 43.1 & 25.8 & 86.6 & \textbf{82.8} & 6.1 & 44.7 & 44.7 \\
                    \bottomrule
                    \end{tabular}
               % \vspace{-0.2cm}
                \end{table*}

            \paragraph{Competitors \& Metrics} 
                The original paper \cite{liu2024visualwebbench} reports the performance of a number of general-purpose and GUI-specific LVLMs. Based on the existing results, we additionally report the performance of Monkey~\cite{li2024monkey}, \edge, and the latest proposed Qwen2-VL \cite{wang2024qwen2}, a general-purpose LVLM but trained with a considerable proportion of GUI grounding data and is therefore classified as semi-GUI-specific. When evaluating the grounding capability of GUI-specific models, we adopt point prediction and report the click accuracy, the proportion of predicted points falling within the ground truth bounding box, rather than the default multiple-choice settings with selection accuracy for general LVLMs. We also slightly modify the input prompt of \edge in the OCR, grounding, and captioning tasks to make them more consistent with the input form of \edge. All other settings remain the same as the original paper \cite{liu2024visualwebbench}.
                
            \paragraph{Results}
                As shown in Table~\ref{tab:visualwebbench}, \edge achieves the best average score among GUI-specific models, and even approximates general-purpose LVLMs with significantly more parameters. First, in the formalized Element OCR task, \edge performs far better than other GUI models and is close to the strongest Claude Sonnet. We specially note the comparison of the results of the Element Ground and Action Ground tasks. According to the original paper \cite{liu2024visualwebbench}, while performing well in the multi-choice setting, general LVLMs can barely handle the more difficult point predictions setting. However, \edge's point prediction performance in the two grounding tasks is not only excellent among GUI models, but also quite competitive with the results of proprietary models in the multi-choice setting. Moreover, for open-ended WebQA and Caption that require local and global semantics, compared with SeeClick, \edge achieves significant gains benefiting from multi-granularity training data. 

                In action ground and other open-ended QA scenarios requiring reasoning skills, \edge's performance is not as good as CogAgent and general LVLMs, which may be due to their larger and more diverse natural language training, while \edge primarily focuses on formalized OCR and grounding tasks.

        \subsubsection{ScreenSpot \cite{cheng2024seeclick}} \label{exp:ss}
            Encompassing over 1200 instructions from 600+ GUI screenshots across web, desktop, and mobile environments, it is specifically designed to evaluate LVLMs' action grounding capabilities and characterized by a substantial number of icon or widget grounding samples.
           
            \paragraph{Competitors \& Metrics} 
                ScreenSpot reports the performance of a general LVLM with grounding capability and GUI-specific LVLMs. Similarly. The evaluation metric is click accuracy as well.
            
            \paragraph{Results} 
                \begin{table*}[ht!] 
                \centering
                \caption{Click accuracy on ScreenSpot, which are sourced from \cite{cheng2024seeclick}, except Qwen2-VL and \edge. $\Circle$ represents a general model, $\CIRCLE$ for a GUI-specific one, and the half-filled $\RIGHTcircle$ represents a general model that specifically trained with GUI grounding data.}
                \label{tab:screenspot}
                \vspace{-0.2cm}
                \begin{tabular}{ccccccccccc}
                \toprule
                 \multirow{2}{*}{LVLMs} & \multirow{2}{*}{\parbox{1.5cm}{\centering Model\\Size}}   & \multirow{2}{*}{\parbox{1.5cm}{\centering GUI\\Specific}} & \multicolumn{2}{c}{Mobile}        & \multicolumn{2}{c}{Desktop} & \multicolumn{2}{c}{Web} & \multirow{2}{*}{Average} \\ \cline{4-9}
                 & & & Text & Icon/Widget & Text & Icon/Widget & Text & Icon/Widget & \\ 
                \midrule
                MiniGPT-v2~\cite{chen2023minigpt} & 7B & \Circle & 8.4  & 6.6  & 6.2 & 2.9 & 6.5 & 3.4 & 5.7 \\
                Qwen-VL~\cite{bai2023qwenvl}    & 9.6B & \Circle & 9.5  & 4.8  & 5.7 & 5.0 & 3.5 & 2.4 & 5.2  \\ 
                Qwen2-VL~\cite{wang2024qwen2}   & 8.3B & \RIGHTcircle & 75.1 & \textbf{60.3} & 73.2 & \textbf{60.7}	& 45.2 & 28.2 & 57.5 \\
                GPT-4V~\cite{achiam2023gpt}     & -    & \Circle & 22.6 & 24.5 & 20.2 & 11.8 &9.2 & 8.8 & 16.2 \\
                CogAgent~\cite{hong2023cogagent}   & 18B  & \CIRCLE & 67.0 & 24.0 & 74.2 & 20.0 & \textbf{70.4} & 28.6 & 47.4 \\
                SeeClick~\cite{cheng2024seeclick}   & 9.6B & \CIRCLE & 78.0 & 52.0 & 72.2 & 30.0 & 55.7 & 32.5 & 53.4 \\ 
                \hline
                \edge & 9.6B & \CIRCLE & \textbf{81.3} & 59.8 & \textbf{75.3} & 43.6 & 68.3 & \textbf{41.7} & \textbf{63.6} \\
                \bottomrule
                \end{tabular}
                \end{table*}
                
                As shown in Table~\ref{tab:screenspot}, compared with general LVLMs, GUI-specific models demonstrate significant improvements. \edge achieved the best average performance across different scenarios, significantly outperforming both SeeClick and CogAgent, demonstrating the effectiveness of our data framework. Additionally, \edge successfully generalized the GUI interacting capabilities to unseen iOS and desktop environments, indicating its transferability. 
            
                In terms icon samples, \edge shows a evident performance improvement compared to SeeClick and CogAgent, underscoring the benefits of our framework in addressing the challenges of collecting icon annotations. The strong competitor, Qwen2-VL, slightly outperforms \edge in mobile and desktop icon samples, likely due to a better model architecture, flexible resolution settings, and a larger, higher-quality set about icon annotations. Unfortunately, Qwen2-VL has not released its data construction details yet.
    
    \subsection{Results on Agent Benchmarks} \label{sec:agent_benchmarks}
        We further explore the benefits of \edge's grounded GUI understanding and interacting on downstream agent benchmarks \cite{shi2017world, rawles2024androidinthewild, deng2024mind2web}. Here each sample typically represents a multi-step interaction environment. Models are required to determine the next action based on overall instruction and current screenshots as well as the operation history, which involves sequential decision-making that goes beyond the grounding and interaction tasks focused on in \edge. Therefore, we highlight the improvements relative to the baselines (Monkey and Seeclick), rather than the absolute results.
        
        We completely follow the experimental setup in SeeClick \cite{cheng2024seeclick}, that is, for each benchmark, we further fine-tune \edge using the exact same respective training split, and evaluate it on the corresponding test set with the same metrics.
        \begin{table}[h]
            \centering
            \caption{Average success rates (SR) on two subset of tasks settings in MiniWob. $\Circle$ means using HTML as input while $\CIRCLE$ means screenshots instead and $\RIGHTcircle$ means using both of them.}
            \label{tab:miniwob}
            \tabcolsep=0.05cm
            \vspace{-0.2cm}
            \begin{tabular}{cccccc}
            \toprule
            Methods & Visual & Subset & \#Tasks & Dataset & Avg.~SR \\
            \midrule
             CC-Net (SL)~\cite{pmlr-v162-humphreys22a} & \RIGHTcircle & 1 & 45 & 2.4M & 35.6 \\
             WebGUM~\cite{furuta2023multimodal} & \RIGHTcircle & 1 & 45 & 2.8K & 65.5 \\
             WebGUM~\cite{furuta2023multimodal} & \RIGHTcircle & 1 & 45 & 347K & \textbf{86.1} \\
             SeeClick~\cite{cheng2024seeclick} & \CIRCLE & 1 & 45 & 2.8K & 73.6 \\
             Monkey~\cite{li2024monkey} & \CIRCLE & 1 & 45 & 2.8K & 77.8 \\
             \edge & \CIRCLE & 1 & 45 & 2.8K & 79.6 \\
             \hdashline
             CC-Net (SL)~\cite{pmlr-v162-humphreys22a} & \CIRCLE & 2 & 35 & 2.4M & 23.4 \\
             Pix2Act~\cite{shaw2023pixels} & \CIRCLE & 2 & 35 & 1.3M & 64.6 \\
             Qwen-VL~\cite{wang2024qwen2} & \CIRCLE & 2 & 35 & 2.8K & 48.4 \\
             SeeClick~\cite{cheng2024seeclick} & \CIRCLE & 2 & 35 & 2.8K & 67.0 \\
             Monkey~\cite{li2024monkey} & \CIRCLE & 2 & 35 & 2.8K & 70.5 \\
             \edge & \CIRCLE & 2 & 35 & 2.8K & \textbf{70.9} \\
            \bottomrule
            \end{tabular}        
        \end{table}

        \begin{table*}[h!]
        \centering
        \caption{Action matching scores of different methods on AITW tasks. ClickAcc is the accuracy of click operations. $\Circle$ represents textual metadata as input while $\CIRCLE$ represents screenshots instead. \dag~indicates the model is fine-tuned on each subset respectively, while other models are unified models across all subsets. Some baselines' results are directly cited from \cite{zhang2023you, hong2023cogagent, yan2023gpt, cheng2024seeclick, rawles2024androidinthewild}.}
        \label{tab:aitw}
        \begin{tabular}{cccccccccc}
        \toprule
        Methods & Visual & General & Install & GoogleApps & Single & WebShopping & Overall & ClickAcc \\
        \midrule 
        LLaMA2-7B$^\dag$\cite{touvron2023llama2openfoundation} & \Circle & 28.6 & 35.2 & 31.0 & 27.4 & 19.9 & 28.4 & & - \\
        PaLM2-CoT~\cite{anil2023palm2technicalreport}$_{(\text{few shot})}$ & \Circle & - & - & - & - & - & 39.6 & - \\
        GPT-3.5~\cite{chagpt}$_{(\text{few shot})}$ & \Circle & 5.9 & 4.4 & 10.5 & 9.4 & 8.4 & 7.7 & - \\
        GPT-4V~\cite{achiam2023gpt}$_{(\text{zero shot})}$ & \CIRCLE & 41.7 & 42.6 & 49.8 & 72.8 & 45.7 & 50.5 & - \\
        Qwen-VL~\cite{bai2023qwenvl} & \CIRCLE & 49.5 & 59.9 & 46.9 & 64.7 & 50.7 & 54.3 & 57.4 \\
        SeeClick~\cite{cheng2024seeclick} & \CIRCLE & 54.0 & 66.4 & 54.9 & 63.5 & 57.6 & 59.3 & 66.4 \\
        Monkey~\cite{li2024monkey} & \CIRCLE & 61.8 & 71.0 & 58.4 & \textbf{75.4} & 65.7 & 66.5 & 65.9 \\
        Auto-UI~\cite{zhang2023you} & \CIRCLE & 71.4 & 76.9 & 70.3 & 68.2 & 84.6 & 74.3 & -\\
        CogAgent~\cite{hong2023cogagent} & \CIRCLE & \textbf{75.0} & \textbf{78.9} & \textbf{71.7} & 65.4 & \textbf{93.5} & \textbf{76.9} & - \\
        \hline
        \edge & \CIRCLE & 66.9 & 73.5 & 62.7 & 74.1 & 68.2 & 69.1 & \textbf{76.0} \\
        \bottomrule
        \end{tabular}
        \end{table*}
        
        \subsubsection{MiniWob \cite{shi2017world}}
            MiniWob is a simplified web environment that includes about 100 types of web automation tasks, where the agent is asked to accomplish open-ended instructions. Our training set contains 2.8k episodes. 

            \paragraph{Competitors \& Metrics} 
                We compare \edge with several offline training methods, including text-based and vision-based. The current state-of-the-art method WebGUM \cite{furuta2023multimodal} use screenshots as auxiliary input but still interact with the environment through HTML elements selecting. Pix2Act \cite{shaw2023pixels} is an vision-based approach trained with extensive demonstration data. Due to the variance in test splits among different methods \cite{liu2018reinforcement, furuta2023multimodal, shaw2023pixels}, for fairness, we report performance in two groups based on the overlapping tasks. We compute the success rate over 50 random seeds for each task and then compute the average over all tasks as the final score.
            \paragraph{Results}
                As reported in Table~\ref{tab:miniwob}, our purely vision-based \edge outperforms the offline SOTA WebGUM \cite{furuta2023multimodal} with HTML and screenshots as input under the same 2.8K training data setting. Benefiting from the powerful LVLM baseline and our grounded GUI training, \edge also surpass vision-based SeeClick \cite{cheng2024seeclick} and Pix2Act \cite{shaw2023pixels}, the latter of which uses nearly 500 times more data.
        
        \subsubsection{Android In The Wild (AITW) \cite{rawles2024androidinthewild}}
            As a smartphone environment benchmark consists of 715k episodes spanning 30k unique instructions, AITW considers a wide range of action types on mobile devices, including tap, swipe, typing, going home, back, etc. 

            \paragraph{Competitors \& Metrics} 
                We compare \edge with two types of models: the methods based on the structured text provided by the original dataset and those visually perceive screenshots. Following SeeClick's \cite{cheng2024seeclick} setup, we adopt the screen-wise action matching score from the original AITW paper \cite{rawles2024androidinthewild} and additionally report the click accuracy (ClickAcc), the accuracy when both reference and prediction are click operations.
            
            \paragraph{Results}
                As displayed in Table~\ref{tab:aitw}, although not as good as Auto-UI and CogAgent, \edge outperforms general LVLMs like GPT-4V and Qwen-VL, as well as SeeClick trained in single-granularity data. These results support our idea that rich multi-granularity grounded training enhances the performance of downstream agents tasks.

        \subsubsection{Mind2Web}
            Mind2Web \cite{deng2024mind2web} comprises over 2000 open-ended tasks collected from 137 real websites, each with high-level instruction for developing and evaluating generalist web agents.

            \paragraph{Competitors \& Metrics} 
                Originally designed for LLM-based agents, the baseline \cite{deng2024mind2web} employs a two-stage approach, where a language model first filter candidate elements from raw HTML, followed by a Flan-T5-XL \cite{chung2024scaling} selecting the target element in a multi-choice format. Other text-based model share the same filtered HTML and multi-choice setup. We follow the setting of \cite{deng2024mind2web} and report the step success rate (step SR) as metric. For LVLMs, we use screenshots instead of HTML as input and report the click accuracy.

            \paragraph{Results}
                \begin{table}[t]
                \centering
                \tabcolsep=0.05cm
                \caption{Step SR on Mind2Web. The column "Task" reports cross-task result, and so on. \dag~indicates that top-10 candidates were used, otherwise top-50. $^*$ indicates the model is fine-tuned, while $^+$ indicates in-context learning examples are given. $\Circle$ means HTML as input while $\CIRCLE$ means screenshots instead. Apart from Monkey, Qwen2-VL and our model, other results are from \cite{deng2024mind2web, gur2023real, lai2024autowebglm, cheng2024seeclick, hong2023cogagent}.}
                \label{tab:mind2web}
                \vspace{-0.3cm}
                \begin{tabular}{ccccccc}
                \toprule
                Methods & \#Params & Visual & Task & Website & Domain & Ave. \\
                \midrule
                Flan-T5-XL$^*$~\cite{deng2024mind2web} & 3B & \Circle & 52.0 & 38.9 & 39.6 & 43.5 \\
                AutoWebGLM~\cite{lai2024autowebglm} & 6B & \Circle & 66.4 & 56.4 & 55.8 & 59.5 \\
                Html-T5-XL$^*$~\cite{gur2023real} & 543B & \Circle & \textbf{71.5} & \textbf{62.2} & \textbf{67.1} & \textbf{66.9} \\
                
                GPT-3.5-Turbo$^+$~\cite{chagpt} & - & \Circle & 17.4 & 16.2 & 18.6 & 17.4 \\
                GPT-4$^{\dag}$~\cite{achiam2023gpt} & - & \Circle & 36.2 & 30.1 & 26.4 & 30.9 \\ 
                
                Qwen-VL$^*$~\cite{bai2023qwenvl} & 9.6B & \CIRCLE & 13.3 & 9.1 & 12.0 & 11.5 \\
                Monkey$^*$~\cite{li2024monkey} & 9.6B & \CIRCLE & 20.3 & 12.0 & 12.2 & 14.8 \\
                SeeClick$^*$~\cite{cheng2024seeclick} & 9.6B & \CIRCLE & 25.5 & 16.4 & 20.8 & 20.9 \\
                CogAgent$^*$~\cite{hong2023cogagent} & 18B & \CIRCLE & 62.3 & 54.0 & 59.4 & 59.5 \\
                \hline
                \edge & 9.6B & \CIRCLE & 30.0 & 21.1 & 22.4 & 24.5 \\
                \bottomrule
                \end{tabular}
                \vspace{-0.3cm}
                \end{table}

                As shown in Table~\ref{tab:mind2web}, \edge offers evident improvements over the base model Monkey and the competitor SeeClick. However, it still lags significantly behind CogAgent and other text-based models, indicating the heavy dependence of real-world web agents on planning capabilities and highlighting the substantial room for improving vision-based agents.

    \subsection{Ablation Study} \label{sec:abl_study}
        % Within the \edge framework, the elementary tasks (i.e., element-level OCR and grounding) are designed to train the model's basic visual and spatial awareness, enhancing its grounded GUI understanding. Building on this foundation, the advanced task setup introduces multi-granularity comprehension and reasoning, further strengthening the model's interaction capabilities. 
        
        To validate the effectiveness of the advanced tasks, we design an experiment where all advanced task samples (i.e., function inference, detailed description, conversation intention) were removed from \edge. The model was then retrained under identical conditions.

        \begin{table}[t]
        \centering
        \caption{Ablation study on advanced tasks. Following the same setup and metrics in Section~\ref{exp:vwb} and Section~\ref{exp:ss}, we report average scores on VisualWebBench and ScreenSpot.}
        \label{tab:abl}
        \vspace{-0.3cm}
        \begin{tabular}{ccc}
        \toprule
        Dataset & VisualWebBench \cite{liu2024visualwebbench} & ScreenSpot \cite{cheng2024seeclick} \\
        \midrule
        w/o Advanced tasks & 43.7 & 62.3 \\
        Full \edge & 44.7 & 63.6 \\
        \bottomrule
        \end{tabular}
         \vspace{-0.3cm}
        \end{table}

\begin{figure}[]
        \centering
        \includegraphics[width=0.75\linewidth]{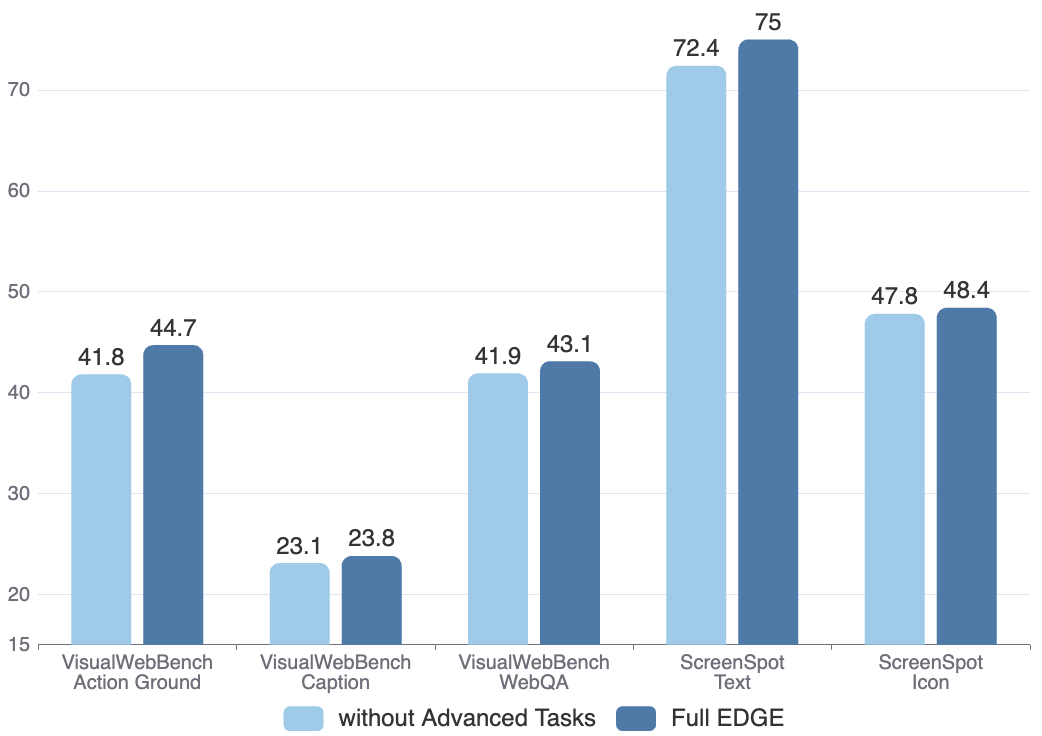}
        \captionsetup{skip=3pt}
        \caption{Ablation results on VisualWebBench and ScreenSpot. Scores of ScreenSpot are averaged over three environments.}
        \Description{The bar chart of the detailed ablation results on VisualWebBench and ScreenSpot. In all tasks, model trained with the full \edge performs better.}
        \label{fig:ablation}
        \end{figure}

        As shown in Table~\ref{tab:abl}, the model trained without advanced tasks performs worse on both GUI benchmarks.  It is important to note that the advanced task samples only constitute 0.5\% of the total dataset, yet their efficiency in further improving the model's capabilities is evident. We also provide a detailed report in Figure~\ref{fig:ablation}, covering global understanding and interaction tasks in both benchmarks. In all these scenarios, the full \edge consistently outperformed its ablated variant. This demonstrates that the training of multi-granularity and reasoning abilities in the advanced tasks enhances the model's GUI understanding and interaction capabilities.

        \begin{figure}[]
        \centering
        \includegraphics[width=\linewidth]{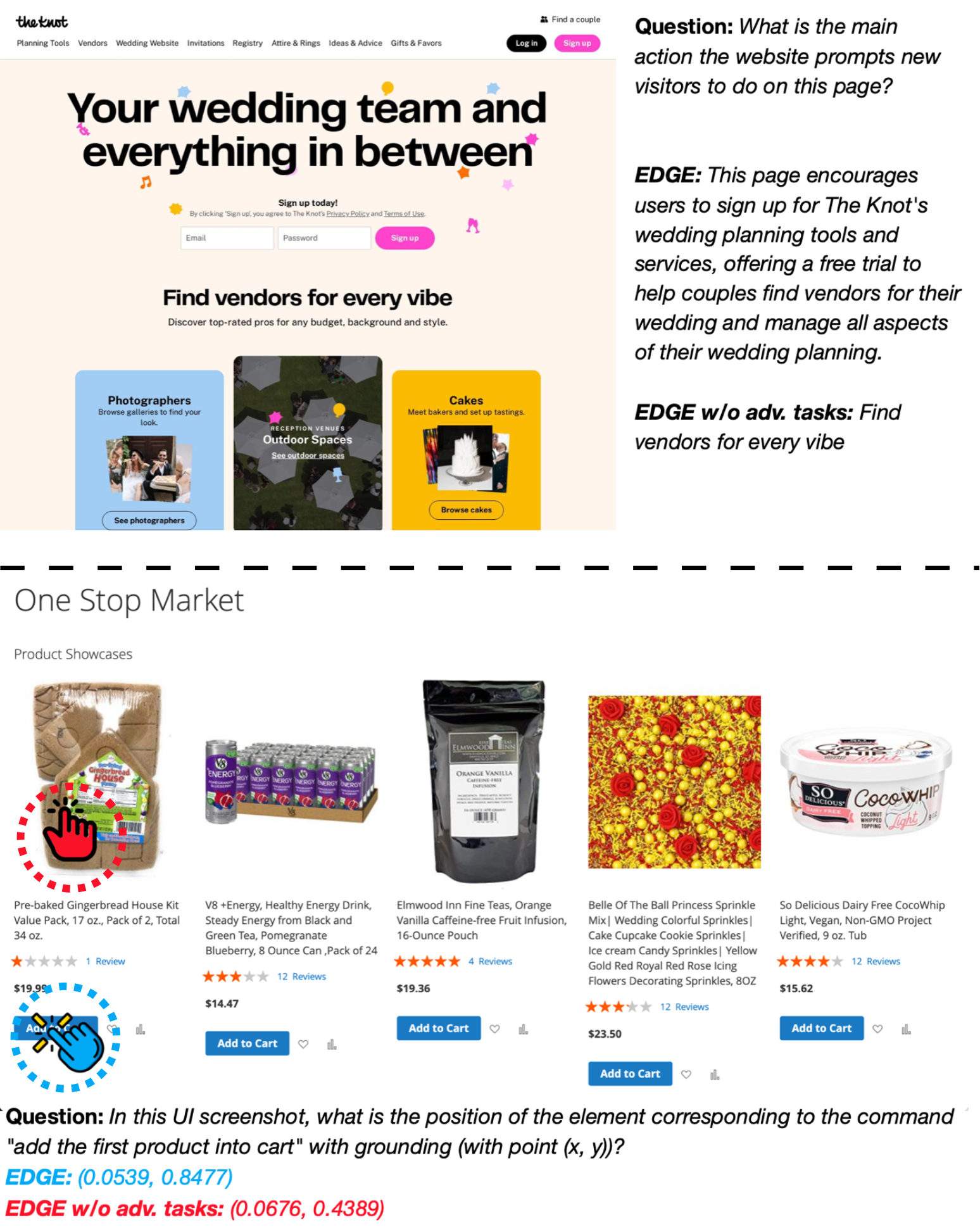}
        \captionsetup{skip=3pt}
        \caption{Examples of comparing the inference results of two models on VisualWebBench and ScreenSpot.}
        \Description{Comparison of the models on two real scenarios. The full \edge correctly understands the global and local level semantics, rather than just repeating the heading or directly clicking on the referenced object.}
        \label{fig:contrast}
        \end{figure}

        We further present two toy examples in Figure~\ref{fig:contrast} to display the different outputs of the full \edge and its ablated variant. The full \edge perfectly answered the questions in the first screenshot, going beyond repeating the heading. In the second one, it successfully distinguished between the product itself and the button for adding it to the cart. These examples provide intuitive demonstrations of how training on advanced tasks can enhance the model. Additional demonstration examples are provided in Appendix~\ref{app:casestudy}.

\section{Limitations and Prospects}
\edge still has significant room for expansion in terms of data scale and diversity. This includes leveraging more web data, introducing additional actions such as typing and double-clicking, and integrating the planning capabilities required for agent tasks.

In the future, the data framework and open-source LVLMs still offer a wealth of topics for researchers to explore. For example, extracting lists and tables, as well as topological relationships between nodes during web annotation, can enhance the learning of structured information extraction. The quantity and quality of icon annotations also need enhancement. Additionally, the instruction-following capabilities of LVLMs are currently weaker compared to similarly sized LLMs, and overcoming this limitation would significantly enhance their potential as truly general agents.
\section{Conclusion}
This paper proposes an \edge, enriched, multi-granularity, and fully automated data synthesis framework for open-source LVLMs, aimed at enhancing their performance on GUI tasks. Based on the annotation of large-scale webpages, we designed elementary recognition tasks and advanced operational tasks, as well as additional icon understanding task, which enable LVLMs to acquire the multi-granularity capabilities required for understanding and interacting with web GUIs, with seamless transfer to mobile and desktop environments. The model's excellent performance in GUI benchmarks and improvements in agent benchmarks validate the effectiveness of \edge. Our work demonstrates that the vast resources available on the Internet hold significant value for various endeavors, including but not limited to LVLMs training. Furthermore, even individual researchers with limited manpower and funding can effectively leverage these resources to advance their work.

\bibliographystyle{ACM-Reference-Format}
\bibliography{references}

\newpage
\appendix

\section{The Annotation Script} \label{app:script}
We meticulously implement a JavaScript script for annotating any webpage, which can be executed directly in the browser console or via browser drivers. It first filters out invisible elements within the current viewport, followed by excluding visually hidden elements, even if present in the HTML, by checking factors like the computed styles and cursor accessibility, as well as the special cases like overflow beyond its container’s boundaries, to ensure accurate visibility judgments. In our experience, this precise visibility filtering is crucial for the accuracy of subsequent element annotations.

\begin{figure}[h!]
    \centering
    \includegraphics[width=\linewidth]{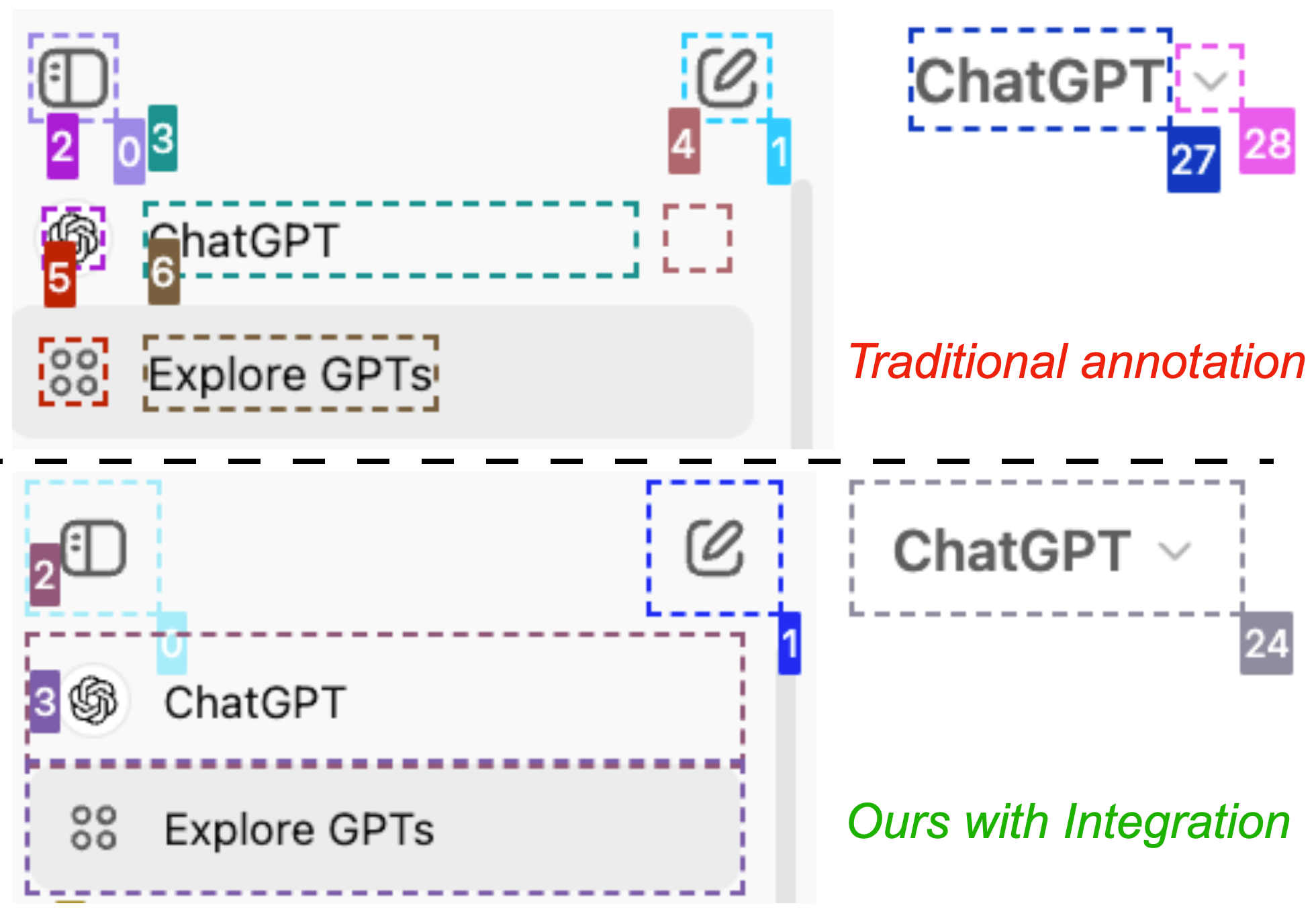}
    \captionsetup{skip=3pt}
    \caption{Comparison of our annotations with traditional methods, where ours capture the complete semantics.}
    \Description{Comparison of our annotations with traditional methods on the top left corner of the page of ChatGPT.}
    \label{fig:app_anno_contrast}
\end{figure}

\begin{figure*}[h!]
    \centering
    \includegraphics[width=\textwidth]{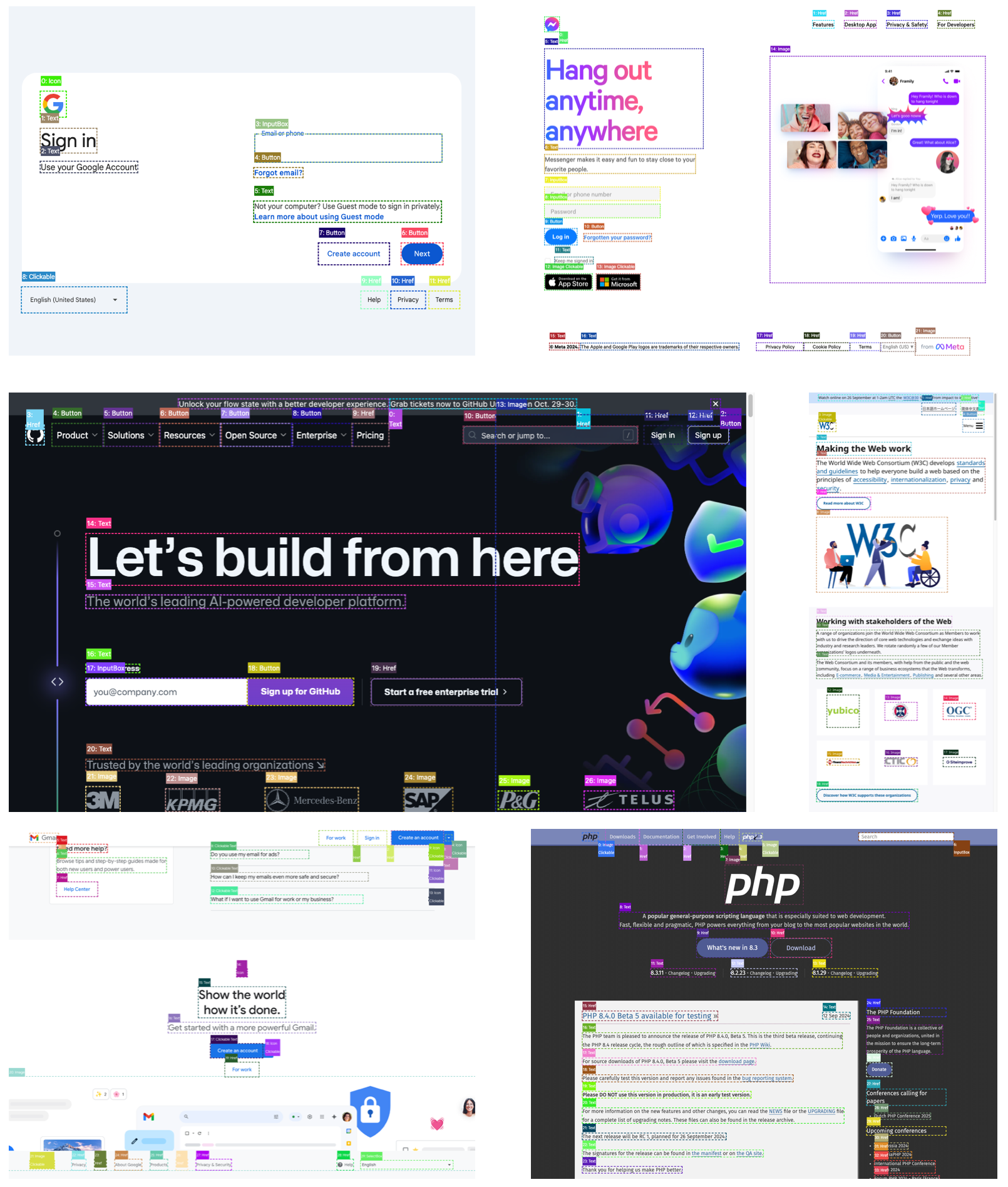}
    \captionsetup{skip=3pt}
    \caption{More webpage annotations results.}
    \Description{More webpage annotations results.}
    \label{fig:app_anno_examples}
\end{figure*}

Next, we determine the type of each visible element based on attributes like tag, role, and computed styles, covering common elements found on webpages, such as text, code, images, icons, buttons, hyperlinks, and inputs. Then we designed a set of integration rules based on the type, content, and children of an element to decide whether it represents the smallest, complete semantic unit visually. For example (illustrated in Figure~\ref{fig:app_anno_contrast}), buttons often consist of borders, text, and icons visually. Traditional methods tend to recognize the text and icon separately, leading to incomplete semantic understanding. In contrast, we treat the entire button as a single element and use the outermost border to determine its position. This approach aligns more closely with human visual perception, providing a more cohesive semantic representation. Figure~\ref{fig:app_anno_examples} shows the more webpage annotations results.

In practice, to avoid samples being concentrated in the header of webpages, we dynamically determine the number and location of annotations based on the length of webpages. Specifically, if the webpage is short and is completely contained in the current viewport, it will be only annotated once; if the webpage is longer, it will be first annotated at the header, and then we scroll to its bottom to annotated it again; if the webpage is very long, it will be annotated once more in the middle.

In implementation, we traverse the DOM tree, recursively collecting information from each element. This method offers great extensibility. For example, we can additionally annotate relationships between elements, such as parent-child or sibling connections, and to identify structured elements like tables and lists. We have published the complete annotation script in our code repository.

\section{Dataset Details} \label{app:dataset}
    \subsection{Training Samples}
        When rendering webpages, we choose from 16 pre-selected common screen sizes to simulate the presentation of webpages on desktop, tablet and mobile devices. 
        
\begin{figure}
            \centering
            \includegraphics[width=\linewidth]{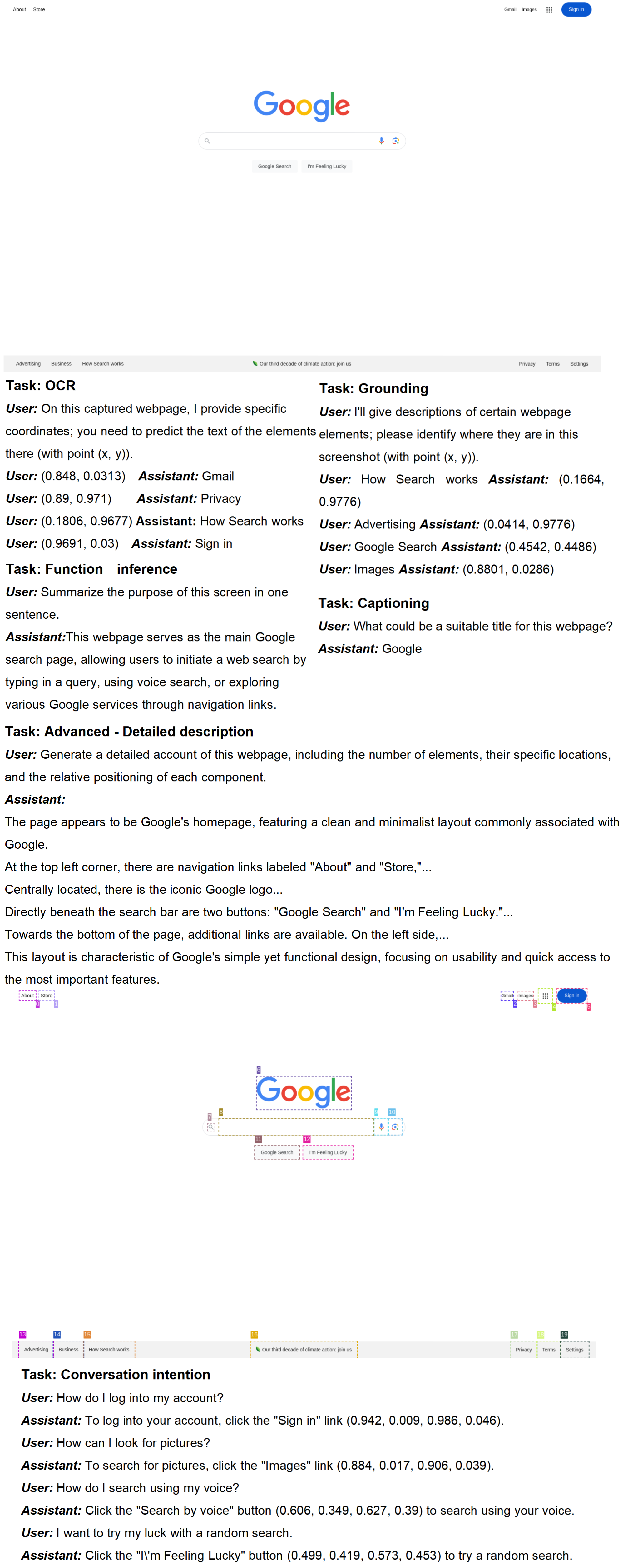}
            \caption{Examples of training samples on the homepage of Google.}
            \Description{Examples of training samples on the homepage of Google.}
            \label{fig:samples}
        \end{figure}

        For elementary tasks, the input images of the training samples are raw screenshots, and the dotted box annotations in the previous images are only for demonstration purposes. We use two question forms in grounding tasks: predicting the bounding box $(x1, y1, x2, y2)$ or its center point $(x, y)$, with the specific choice explicitly indicated in the input. As shown in Figure\ref{fig:samples}, for element-level tasks, we use a multi-turn conversation format, where a sample consists of which is not only more sample efficient, but also bring better experimental results. Page-level tasks still occupy a single sample.

        Since the encoding length of the Monkey~\cite{li2024monkey} for an image exceeds half of the total sequence length, each training sample contains at most one image. We adopt different sample formats for element-level tasks and page-level tasks. As shown in the Figure ~\ref{fig:samples}, for element-level tasks like OCR and grounding, we use a multi-turn conversation format, where each training sample contains multiple independent question-answering instances of a single screenshot on the same task. This setting brings a higher density of supervision signals, makes full use of the calculation of each forward propagation, and can improve the efficiency of model training. Page-level tasks like captioning still occupy a single sample. Page-level question-answering tasks such as captioning still take a single sample. Advanced tasks share a similar division, with element-level the conversation intention task using a multi-turn conversation setting, while the page-level detailed description and function inference use a single question-answer format.

        % \begin{figure}
        %     \centering
        %     \includegraphics[width=\linewidth]{figures/app_vwb_prompt.png}
        %     \captionsetup{skip=3pt}
        %     \caption{Modified task prompts in VisualWebBench.}
        %     \Description{Modified task prompts in VisualWebBench.}
        %     \label{fig:app_vwb_prompt}
        % \end{figure}

    \subsection{Question Templates for Elementary Tasks}
\begin{figure*}
        \centering
        \includegraphics[width=0.92\textwidth]{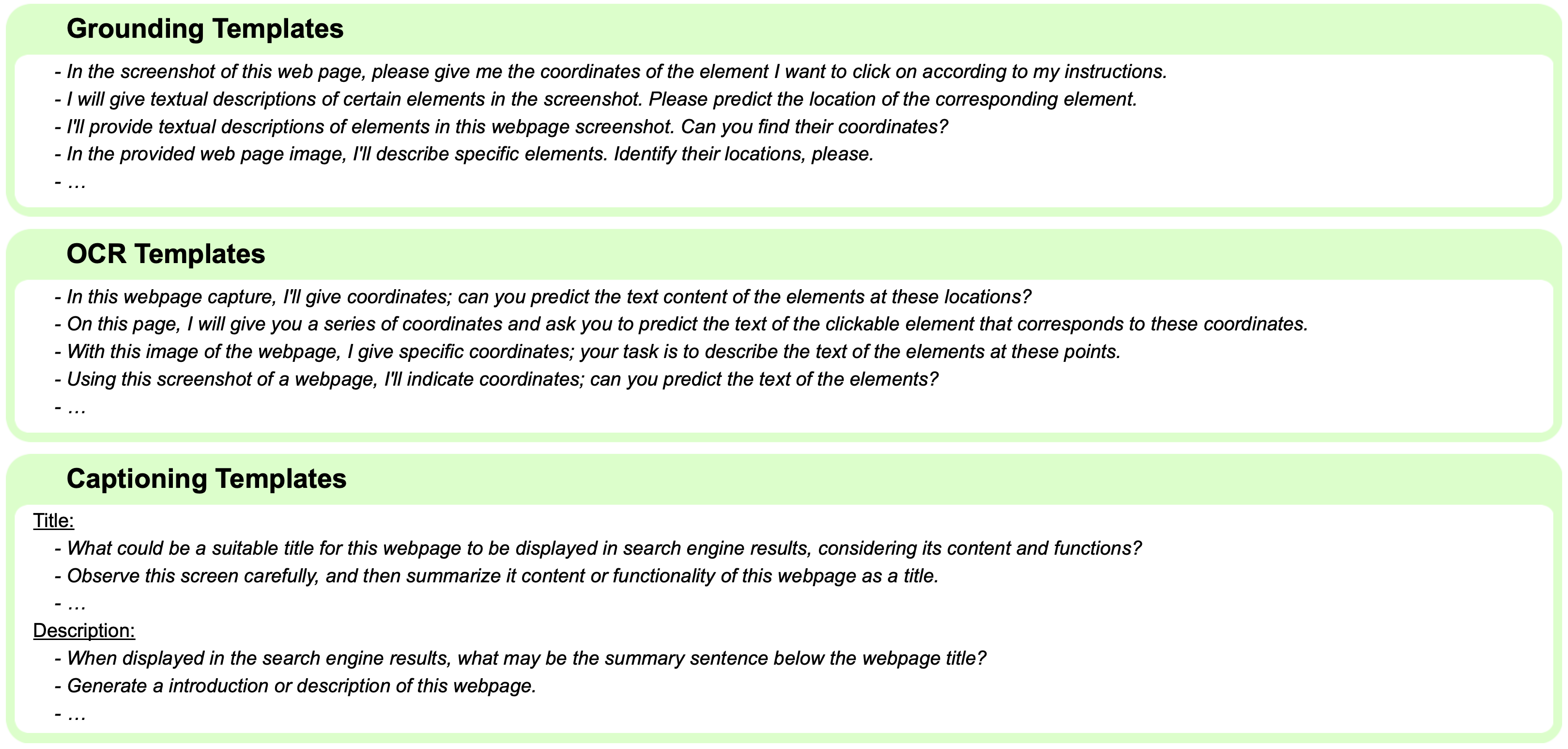}
        \captionsetup{skip=3pt}
        \caption{Question templates for some elementary tasks.}
        \Description{Question templates for some elementary tasks.}
        \label{fig:templates}
    \end{figure*}

        For single-round QA tasks, we first manually write three question templates for each task, and then prompted ChatGPT to generate more templates that are richer and more various in natural language features such as vocabulary, syntax, and sentiment, to simulate the query input by various users with different prompting habits. For multi-turn QA tasks such as OCR and grounding, as well as all of mobile screen tasks on the Rico dataset, we directly use the multi-turn question templates provided by SeeClick~\cite{cheng2024seeclick}, which typically have dozens of templates for each task. Note that as shown in Figure~\ref{fig:samples} in the element-level multi-turn conversation, the "template" we refer to here is the task description. For example, for the grounding task, we first input the task description, followed by multiple elementary questions and answers, i.e., the description and the location of the on-screen elements. As for the page-level tasks, templates are the sepcific questions. Figure~\ref{fig:templates} shows some templates for grounding, OCR, captioning, and icon describing tasks.

    \subsection{Prompts for Advanced Tasks}
        We use Claude-3.5-Sonnet-20240620 to generate advanced task QA data. According to preliminary manual judgment, the quality of QA pairs generated by Claude-3.5 is slightly higher than those by GPT-4V. Following Ferret-UI~\cite{you2024ferret}, the textual prompts we input into Claude include shared prompts, task prompts, screen annotations, and QA examples (only for conversation intention). On this basis, webpage screenshots annotated with the Set-of-Mark~\cite{yang2023set} visual prompting method are used as visual input to help the model capture information and semantics that the automatic annotation may miss. Human evaluation shows that the Set-of-Mark prompted screenshots input effectively utilizes the powerful visual understanding ability of proprietary models, which can not only correct the semantics that are omitted or mislabeled by the annotation script, but also reduce the model's hallucinations. Because we require the model to output the annotation number corresponding to each element, which forces it to select the annotated elements, thereby using accurate positions instead of the inaccurate self-generated ones. 
        
\begin{figure*}
        \centering
        \includegraphics[width=0.92\textwidth]{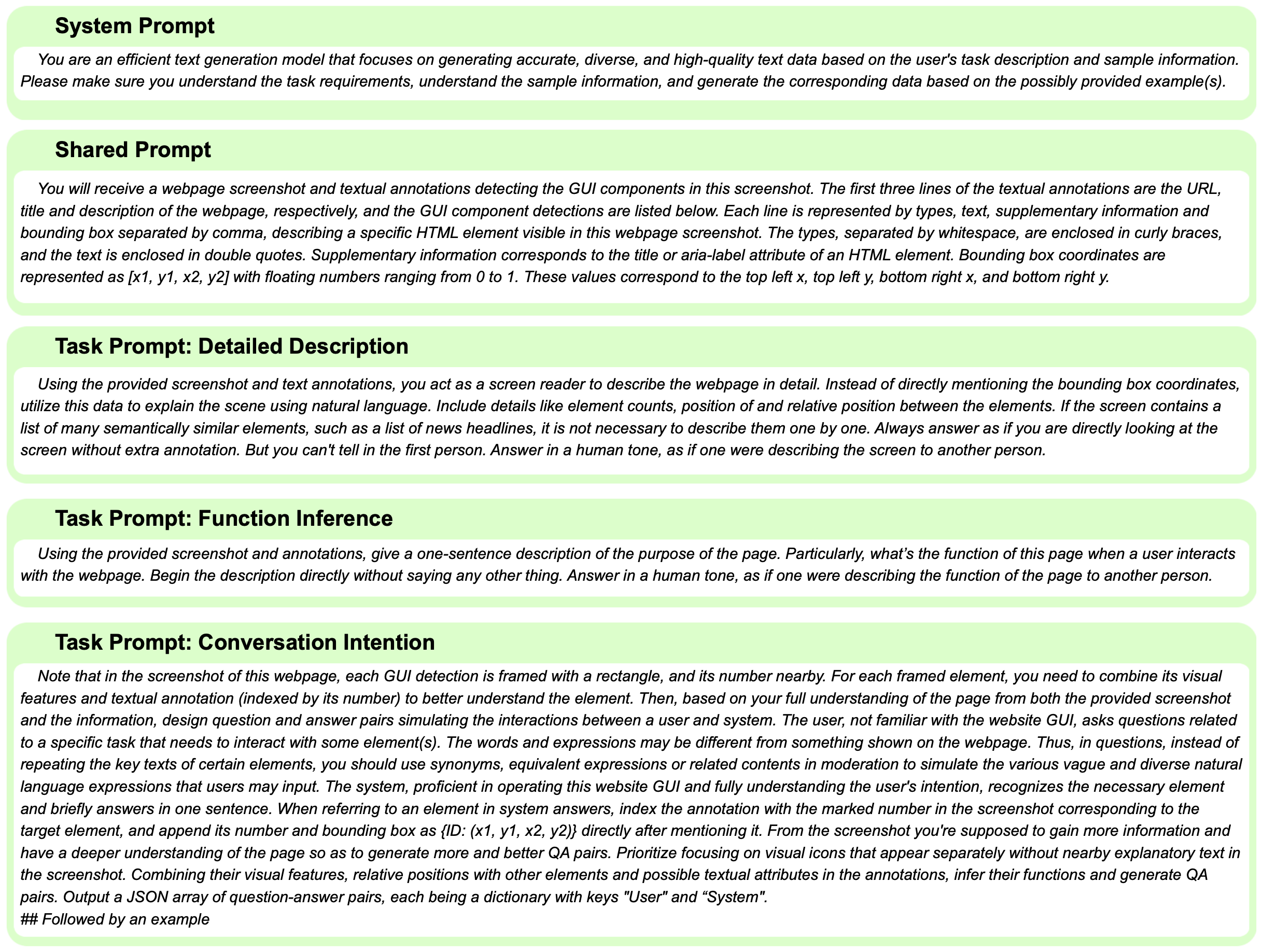}
        \captionsetup{skip=3pt}
        \caption{Prompts for advanced tasks data synthesizing using Claude-3.5. Some prompts are slightly simplified, retaining the core logic and omitting the cumbersome details.}
        \Description{Prompts for advanced tasks data synthesizing using Claude-3.5. Some prompts are slightly simplified, retaining the core logic and omitting the cumbersome details.}
        \label{fig:prompts}
    \end{figure*}

        An example of the conversation intention task is displayed in Figure~\ref{fig:samples}. Examples of the shared prompt and the task prompt for the three advanced tasks are shown in Figure~\ref{fig:prompts}.

\section{Evaluation Details}

    \subsection{VisualWebBench}
        There are seven tasks in VisualWebBench~\cite{liu2024visualwebbench}: Caption, WebQA, HeadOCR, Element OCR, Element Ground, Action Prediction, and Action Ground. 
        
        The Caption task requires models to predict the \texttt{content} attribute of the tag \texttt{<meta name="description" content="xxx">} in the HTML header. The WebQA task involves questions and answers about various contents on webpages. HeadOCR requires the model to output the content of the most eye-catching heading on the webpage. The Element OCR task gives the bounding box of an element and marks the element with a red rectangular box in the screenshot, requiring the model to extract the text content of the element. Element Ground provides the text of an element, marks multiple elements with numbered red rectangular box in the screenshot, and requires the model to select the correct number, that is, the number corresponding to the element containing the given text. In the Action Prediction task, the model choose the most likely \texttt{description} of the webpage that will be jumped to after clicking on an element. In the Action Ground task, the model selects the element to be clicked based on the given user instructions and the screenshot with multiple elements marked with red boxes, also in the form of multiple choices.
    
        When evaluating the GUI understanding and interaction capabilities of LVLMs on VisualWebBench~\cite{liu2024visualwebbench}, we adopt the prompts provided by VisualWebBench by default. The exceptions are that for the evaluation of \edge on the Element Ground and Action Ground tasks, we use our own prompts, requiring the model to directly predict the coordinate point instead of the default multiple choice setting. 
        
        We also slightly modified the prompt of the Caption task to describe the target of the task in a more humane way. In contrast, the original prompt directly requires the model to output \texttt{<meta name="description" content="YOUR ANSWER">}, which is too formalized. Intuitively, neither humans nor GUI interaction models need to understand the HTML structure and content behind the webpage.

\begin{figure}
            \centering
            \includegraphics[width=\linewidth]{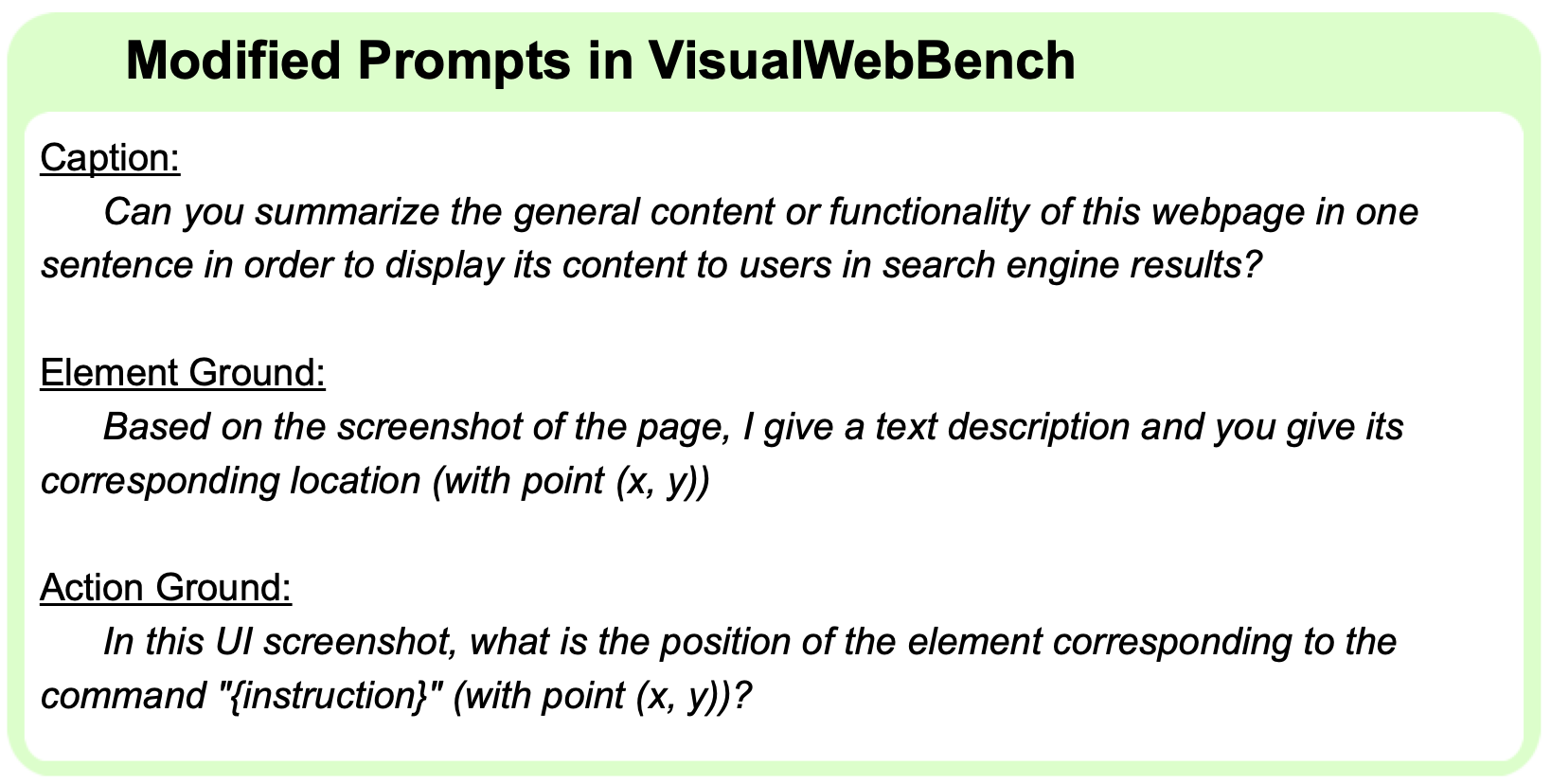}
            \captionsetup{skip=3pt}
            \caption{Modified task prompts in VisualWebBench.}
            \Description{Modified task prompts in VisualWebBench.}
            \label{fig:app_vwb_prompt}
        \end{figure}

        For the above three tasks, our modified prompts are shown in Figure~\ref{fig:app_vwb_prompt}.

    \subsection{ScreenSpot}
        Each sample in ScreenSpot~\cite{cheng2024seeclick} is an action grounding task. Experimental results show that the point prediction of each model is slightly better than the bounding box prediction. Therefore, we use the same prompt as the Action Ground task of VisualWebBench, as shown in Figure~\ref{fig:app_vwb_prompt}.

    \subsection{Downstream Agent Benchmarks}
\begin{figure}
            \centering
            \includegraphics[width=\linewidth]{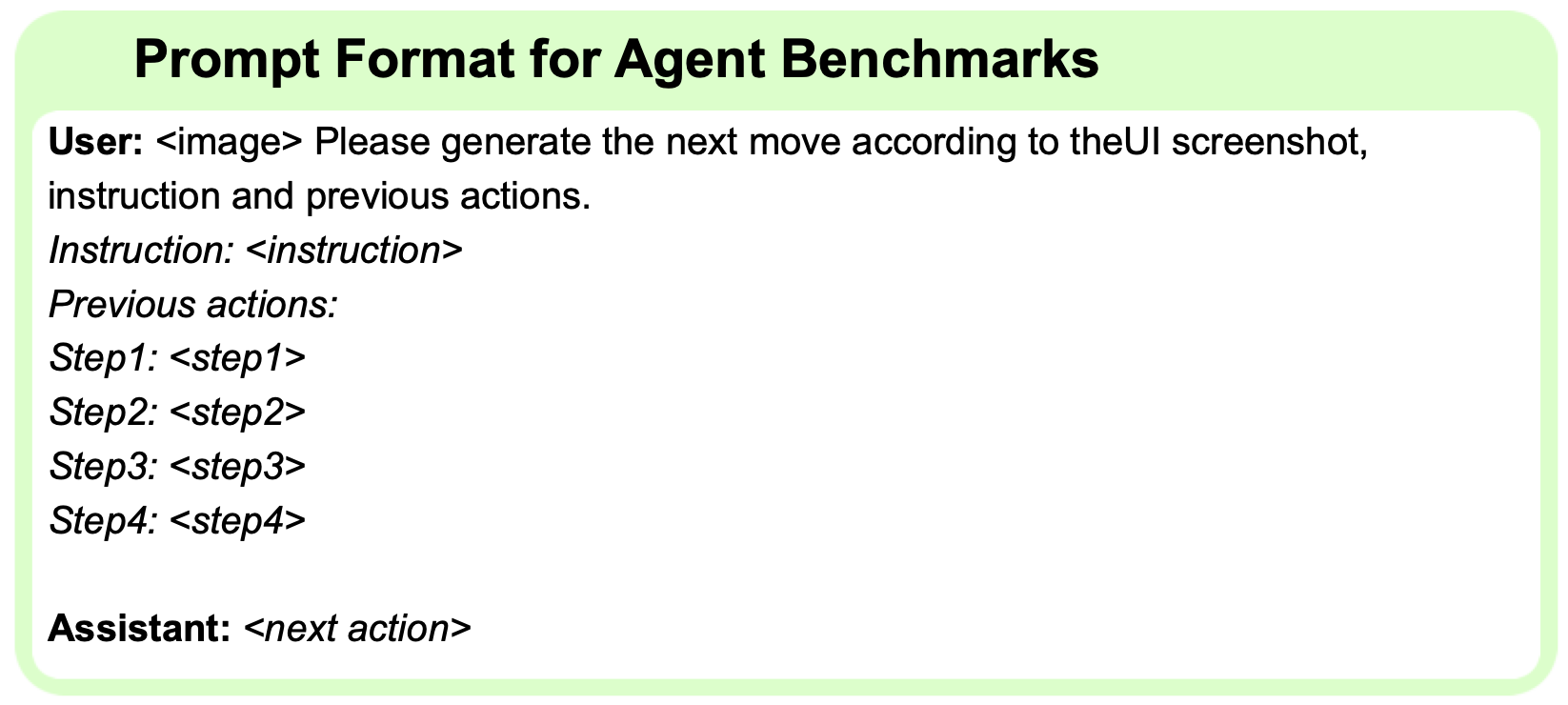}
            \caption{The prompt format for fine-tuning and evaluation in the three downstream agent benchmarks.}
            \Description{The prompt format for fine-tuning and evaluation in the three downstream agent benchmarks.}
            \label{fig:app_agent_prompt}
        \end{figure}

        We use the same dataset and evaluation settings as in SeeClick~\cite{cheng2024seeclick} for fine-tuning and evaluating \edge on the three downstream agent benchmarks, namely MiniWob~\cite{shi2017world}, AITW~\cite{rawles2024androidinthewild}, and Mind2Web~\cite{deng2024mind2web}. Specifically, \edge does not use any explicit agent framework, but simply concatenates instructions and the previous actions as text prompts, and then directly generates the next action. SeeClick has defined a set of mappings from textual actions to numerical action \texttt{ids}, and the fine-tuning process is to make the model familiar with the mapping between the action \texttt{ids} of the corresponding benchmark and the real action in natural language. Figure~\ref{fig:app_agent_prompt} shows the prompt format used by \edge in the fine-tuning and evaluation of these three agent benchmarks, which is exactly the same as that shown in the appendix of the SeeClick paper~\cite{cheng2024seeclick}.

\section{More Inference Results} \label{app:casestudy}
\begin{figure*}
        \centering
        \includegraphics[width=0.85\textwidth]{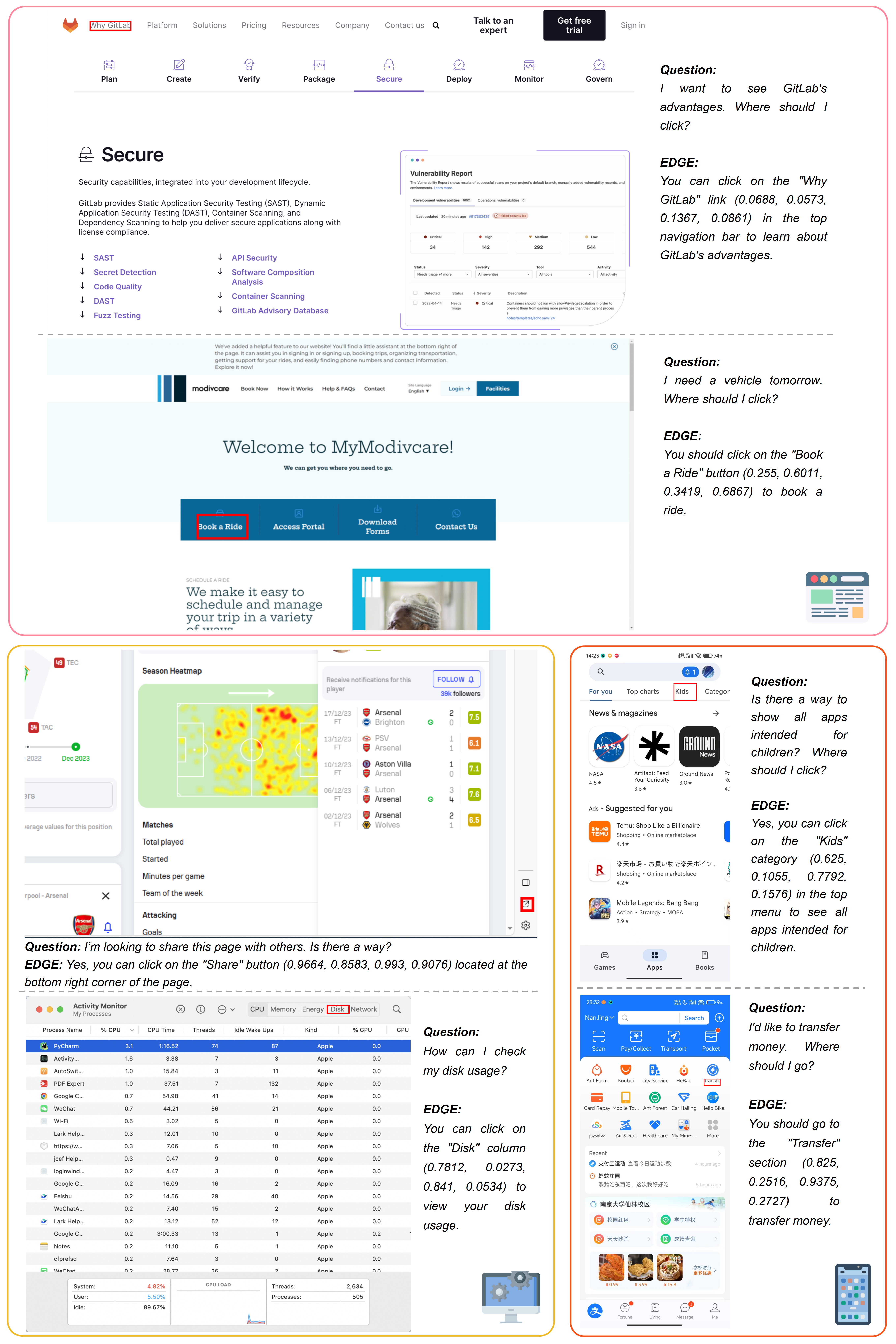}
        \caption{Examples of inference results of \edge on web, desktop, and mobile environments are shown. The output bounding box is marked with a red rectangle.}
        \Description{Examples of inference results of \edge on web, desktop, and mobile environments are shown. The output bounding box is marked with a red rectangle.}
        \label{fig:app_examples}
    \end{figure*}

    More reasoning results are shown in Figure~\ref{fig:app_examples}. Here we ask the model to output the predicted bounding box of the target element instead of a single point to improve interpretability. The output bounding boxes are marked with red rectangular boxes in the screenshots.
    
    The template-based elementary tasks enables the model to accurately locate elements, while after training on free-form high-level tasks, our model supports understanding and answering user operation instructions in a natural, human-like manner. With the help of LVLM's native natural language understanding and generation capabilities, \edge implicitly completes the process of intent recognition, action grounding, and element grounding, just like a human assistant talking to you.
    
\end{document}